\newif\ifJOURNAL
\JOURNALfalse
\newif\ifarXiv
\arXivfalse
\newif\ifFULL
\FULLfalse

\arXivtrue

\ifJOURNAL
  \documentclass[twoside,11pt]{article}
  \usepackage{jmlr2e}
  \usepackage{amsmath,amsfonts,latexsym,algorithm,algorithmic}
\fi

\ifarXiv
  \documentclass{article}
  \usepackage{amsmath,amsthm,amsfonts,amssymb,latexsym,graphicx,algorithm,algorithmic}
  \newcommand{\citealt}[1]{\cite{#1}}
  \newcommand{\citep}[1]{\cite{#1}}
  \newcommand{\citet}[1]{\cite{#1}}
  \newcommand{\citeyear}[1]{\cite{#1}}
\fi

\ifFULL
\usepackage{color}

\newcommand{\bluebegin}{\begingroup\color{blue}}
\newcommand{\blueend}{\endgroup}

\fi

\newlength{\picturewidth}
\ifJOURNAL
\fi
\ifarXiv
\fi

\allowdisplaybreaks[4]

\newcommand{\Vladimir}{Vladimir}
\newcommand{\DOT}{.}
\newcommand{\zzrelax}[1]{}

\newcommand{\T}{^{\text{{\rm T}}}}

\newcommand{\st}{\mathrel{\!|\!}}

\newcommand{\GGG}{\mathcal{G}}				
\newcommand{\PPP}{\mathcal{P}}				

\newcommand{\bbbr}{\mathbb{R}}		

\ifarXiv
\theoremstyle{plain}

\newtheorem{proposition}{Proposition}

\newtheorem{theorem}{Theorem}

\theoremstyle{definition}

\fi

\begin{document}
\title{Prediction with expert advice for the Brier game}

\ifarXiv
  \author{Vladimir Vovk and Fedor Zhdanov\\[2mm]
    Computer Learning Research Centre\\
    Department of Computer Science\\
    Royal Holloway, University of London\\
    Egham, Surrey TW20 0EX, England}
\fi

\maketitle
\begin{abstract}%
  We show that the Brier game of prediction is mixable
  and find the optimal learning rate and substitution function for it.
  The resulting prediction algorithm is applied to predict results of football and tennis matches.
  The theoretical performance guarantee
  turns out to be rather tight on these data sets,
  especially in the case of the more extensive tennis data.
\end{abstract}

\section{Introduction}

The paradigm of prediction with expert advice
was introduced in the late 1980s
(see, e.g., \citealt{desantis/etal:1988}, \citealt{littlestone/warmuth:1994},
\citealt{cesabianchi/etal:1997})
and has been applied to various loss functions;
see \citet{cesabianchi/lugosi:2006} for a recent book-length review.
An especially important class of loss functions
is that of ``mixable'' ones,
for which the learner's loss can be made
as small as the best expert's loss plus a constant
(depending on the number of experts).
It is known \citep{haussler/etal:1998,vovk:1998game}
that the optimal additive constant is attained by the ``strong aggregating algorithm''
proposed in \citet{vovk:1990}
(we use the adjective ``strong'' to distinguish it
from the ``weak aggregating algorithm'' of \citealt{kalnishkan/vyugin:2005}).

There are several important loss functions
that have been shown to be mixable
and for which the optimal additive constant has been found.
The prime examples in the case of binary observations
are the log loss function and the square loss function.
The log loss function,
whose mixability is obvious,
has been explored extensively,
along with its important generalizations,
the Kullback--Leibler divergence and Cover's loss function.

In this paper we concentrate on the square loss function.
In the binary case,
its mixability was demonstrated in \citet{vovk:1990}.
There are two natural directions in which this result
could be generalized:
\begin{description}
\item[Regression:]
  observations are real numbers
  (square-loss regression is a standard problem in statistics).
\item[Classification:]
  observations take values in a finite set
  (this leads to the ``Brier game'', to be defined below,
  a standard way of measuring the quality of predictions in meteorology
  and other applied fields:
  see, e.g., \citealt{dawid:1986}).
\end{description}
The mixability of the square loss function
in the case of observations belonging to a bounded interval of real numbers
was demonstrated in \citet{haussler/etal:1998};
Haussler et al.'s algorithm was simplified in \citet{vovk:2001competitive}.
Surprisingly, the case of square-loss non-binary classification
has never been analysed in the framework of prediction with expert advice.
The purpose of this paper is to fill this gap.
Its short conference version \citep{vovk/zhdanov:2008ICML}
appeared in the ICML 2008 proceedings.

\section{Prediction algorithm and loss bound}
\label{sec:main}

A game of prediction consists of three components:
the observation space $\Omega$,
the decision space $\Gamma$,
and the loss function $\lambda:\Omega\times\Gamma\to\bbbr$.
In this paper we are interested in the following \emph{Brier game}
\citep{brier:1950}:
$\Omega$ is a finite and non-empty set,
$\Gamma:=\PPP(\Omega)$ is the set of all probability measures on $\Omega$,
and
\begin{equation*}
  \lambda(\omega,\gamma)
  =
  \sum_{o\in\Omega}
  \left(
    \gamma\{o\} - \delta_{\omega}\{o\}
  \right)^2,
\end{equation*}
where $\delta_{\omega}\in\PPP(\Omega)$ is the probability measure
concentrated at $\omega$:
$\delta_{\omega}\{\omega\}=1$
and $\delta_{\omega}\{o\}=0$ for $o\ne\omega$.
(For example,
if $\Omega=\{1,2,3\}$, $\omega=1$,
$\gamma\{1\}=1/2$, $\gamma\{2\}=1/4$, and $\gamma\{3\}=1/4$,
$\lambda(\omega,\gamma)=(1/2-1)^2+(1/4-0)^2+(1/4-0)^2=3/8$.)

The game of prediction is being played repeatedly
by a learner having access to decisions made by a pool of experts,
which leads to the following prediction protocol:

\makeatletter
  \renewcommand{\ALG@name}{Protocol}
\makeatother
\begin{algorithm}[H]
  \caption{Prediction with expert advice}
  \label{prot:PEA}
  \begin{algorithmic}
    \STATE $L_0:=0$.
    \STATE $L_0^k:=0$, $k=1,\ldots,K$.
    \FOR{$N=1,2,\dots$}
      \STATE Expert $k$ announces $\gamma_N^k\in\Gamma$, $k=1,\ldots,K$.
      \STATE Learner announces $\gamma_N\in\Gamma$.
      \STATE Reality announces $\omega_N\in\Omega$.
      \STATE $L_N:=L_{N-1}+\lambda(\omega_N,\gamma_N)$.
      \STATE $L_N^k:=L_{N-1}^k+\lambda(\omega_N,\gamma_N^k)$, $k=1,\ldots,K$.
    \ENDFOR
  \end{algorithmic}
\end{algorithm}
\makeatletter
  \renewcommand{\ALG@name}{Algorithm}
\makeatother

At each step of Protocol \ref{prot:PEA} Learner is given
$K$ experts' advice
and is required to come up with his own decision;
$L_N$ is his cumulative loss over the first $N$ steps,
and $L_N^k$ is the $k$th expert's cumulative loss over the first $N$ steps.
In the case of the Brier game,
the decisions are probability forecasts for the next observation.

An optimal (in the sense of Theorem \ref{thm:main} below)
strategy for Learner in prediction with expert advice for the Brier game
is given by the strong aggregating algorithm.
For each expert $k$,
the algorithm maintains its weight $w^k$,
constantly slashing the weights of less successful experts.
Its description uses the notation $t^+:=\max(t,0)$.

\addtocounter{algorithm}{-1}
\begin{algorithm}[ht]
  \caption{Strong aggregating algorithm for the Brier game}
  \label{alg:SAA}
  \begin{algorithmic}
    \STATE $w_0^k:=1$, $k=1,\ldots,K$.
    \FOR{$N=1,2,\dots$}
      \STATE Read the Experts' predictions $\gamma^k_N$, $k=1,\ldots,K$.
      \STATE Set $G_N(\omega):=-\ln\sum_{k=1}^K w^k_{N-1} e^{-\lambda(\omega,\gamma_N^k)}$, $\omega\in\Omega$.
      \STATE Solve $\sum_{\omega\in\Omega}(s-G_N(\omega))^+=2$ in $s\in\bbbr$.
      \STATE Set $\gamma_N\{\omega\}:=(s-G_N(\omega))^+/2$, $\omega\in\Omega$.
      \STATE Output prediction $\gamma_N\in\PPP(\Omega)$.
      \STATE Read observation $\omega_N$.
      \STATE $w_N^k:=w_{N-1}^ke^{-\lambda(\omega_N,\gamma_N^k)}$.
    \ENDFOR
  \end{algorithmic}
\end{algorithm}

The algorithm will be derived in Section \ref{sec:algorithm}.
The following result (to be proved in Section \ref{sec:proof})
gives a performance guarantee for it
that cannot be improved by any other prediction algorithm.

\begin{theorem}\label{thm:main}
  Using Algorithm~\ref{alg:SAA} as Learner's strategy
  in Protocol \ref{prot:PEA} for the Brier game
  guarantees that
  \begin{equation}\label{eq:can}
    L_N
    \le
    \min_{k=1,\ldots,K}
    L_N^k
    +
    \ln K
  \end{equation}
  for all $N=1,2,\ldots$\,.
  If $A<\ln K$,
  Learner does not have a strategy guaranteeing
  \begin{equation}\label{eq:cannot}
    L_N
    \le
    \min_{k=1,\ldots,K}
    L_N^k
    +
    A
  \end{equation}
  for all $N=1,2,\ldots$\,.
\end{theorem}

\noindent
The second part of this theorem follows from its special case
with $\lvert\Omega\rvert=2$ (the binary case).
However, we are not aware of a proof of this result in the binary case,
and we will not use this reduction.

\section{Experimental results}
\label{sec:experimental}

In our first empirical study of Algorithm \ref{alg:SAA}
we use historical data about 6473
matches
in various English football league competitions,
namely:
the Premier League
(the pinnacle of the English football system),
the Football League Championship,
Football League One,
Football League Two,
the Football Conference.
Our data, provided by Football-Data, cover three seasons,
2005/2006, 2006/2007, and 2007/2008.
(The 2007/2008 season ended in May shortly after the ICML 2008 submission deadline,
and so the data set used in the conference version \citep{vovk/zhdanov:2008ICML} of this paper
covered only part of that season,
with 6416 matches in total.)
The matches are sorted first by date, then by league, and then by the name of the home team.
In the terminology of our prediction protocol,
the outcome of each match is the observation,
taking one of three possible values,
``home win'', ``draw'', or ``away win'';
we will encode the possible values as 1, 2, and 3.

For each match we have forecasts made by a range of bookmakers.
We chose eight bookmakers
for which we have enough data over a long period of time,
namely
Bet365, Bet\&Win, Gamebookers, Interwetten,
Ladbrokes, Sportingbet, Stan James, and VC Bet.
(And the seasons mentioned above
were chosen because the forecasts of these bookmakers
are available for them.)

A probability forecast for the next observation
is essentially a vector $(p_1,p_2,p_3)$ consisting of positive numbers summing to $1$.
The bookmakers do not announce these numbers directly;
instead, they quote three betting odds,
$a_1$, $a_2$, and $a_3$.
Each number $a_i$ is the amount
which the bookmaker undertakes to pay out to a client
betting on outcome $i$
per unit stake
in the event that $i$ happens
(the stake itself is never returned to the bettor,
which makes all betting odds greater than $1$;
i.e., the odds are announced according to the ``continental''
rather than ``traditional'' system).
The inverse value $1/a_i$, $i\in\{1,2,3\}$,
can be interpreted as the bookmaker's quoted probability
for the observation $i$.
The bookmaker's quoted probabilities are usually slightly
(because of the competition with other bookmakers)
in his favour:
the sum $1/a_1+1/a_2+1/a_3$ exceeds $1$
by the amount called the \emph{overround}
(at most $0.15$ in the vast majority of cases).
We used
\begin{equation}\label{eq:p}
  p_i := \frac{1/a_i}{1/a_1+1/a_2+1/a_3},
  \quad
  i=1,2,3,
\end{equation}
as the bookmaker's forecasts;
it is clear that $p_1+p_2+p_3=1$.

\begin{figure}[ht]
\begin{center}
\centerline{\includegraphics[width=\picturewidth]{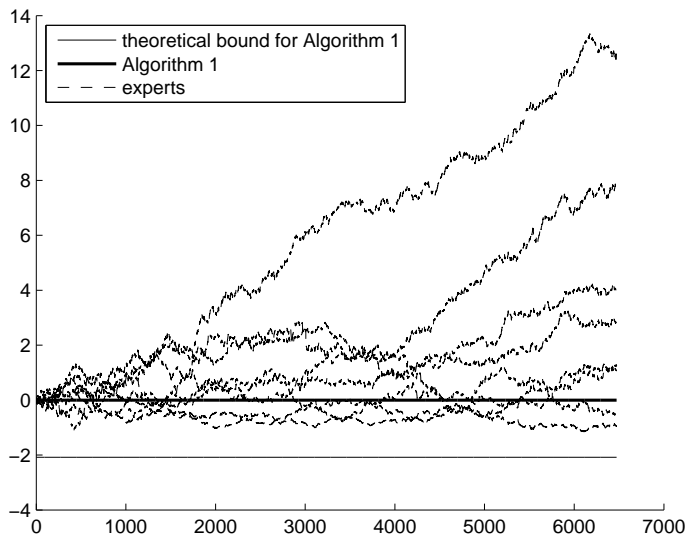}}
\caption{The difference between the cumulative loss of each of the $8$ bookmakers
  (experts)
  and of Algorithm~\ref{alg:SAA}
  on the football data.
  The theoretical lower bound $-\ln 8$ from Theorem~\ref{thm:main} is also shown.}
\label{fig:football}
\end{center}
\end{figure}

The results of applying Algorithm~\ref{alg:SAA} to the football data,
with $8$ experts and $3$ possible observations,
are shown in Figure~\ref{fig:football}.
Let $L_N^k$ be the cumulative loss of Expert $k$, $k=1,\ldots,8$,
over the first $N$ matches
and $L_N$ be the corresponding number for Algorithm~\ref{alg:SAA}
(i.e., we essentially continue to use the notation of Theorem \ref{thm:main}).
The dashed line corresponding to Expert $k$
shows the excess loss
$N\mapsto L_N^k-L_N$ of Expert $k$ over Algorithm~\ref{alg:SAA}.
The excess loss can be negative,
but from Theorem~\ref{thm:main} we know that it cannot be less than $-\ln 8$;
this lower bound is also shown in Figure~\ref{fig:football}.
Finally, the thick line (the positive part of the $x$ axis)
is drawn for comparison:
this is the excess loss of Algorithm~\ref{alg:SAA} over itself.
We can see that at each moment in time
the algorithm's cumulative loss is fairly close
to the cumulative loss of the best expert
(at that time; the best expert keeps changing over time).

\begin{figure}[ht]
\begin{center}
\centerline{\includegraphics[width=\picturewidth]{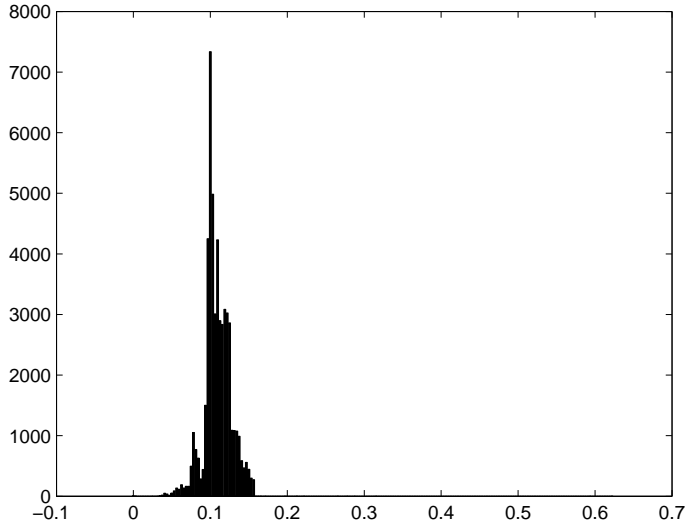}}
\caption{The overround distribution histogram for the football data,
  with $200$ bins of equal size between the minimum and maximum values of the overround.}
\label{fig:football-overround}
\end{center}
\end{figure}

Figure~\ref{fig:football-overround} shows the distribution
of the bookmakers' overrounds.
We can see that in most cases overrounds are between $0.05$ and $0.15$,
but there are also occasional extreme values,
near zero or in excess of $0.3$.
In Figure \ref{fig:football}
one bookmaker clearly performs worse than the others.
His poor performance may be explained by his mean overround being about 0.13,
near the top end of the distribution in Figure~\ref{fig:football-overround}.
(On one hand, a high overround diminishes the need for accurate probability forecasts,
and on the other,
our estimates (\ref{eq:p}) 
of the probabilities implicit in the announced odds
also become less precise.)

\begin{figure}[ht]
\begin{center}
\centerline{\includegraphics[width=\picturewidth]{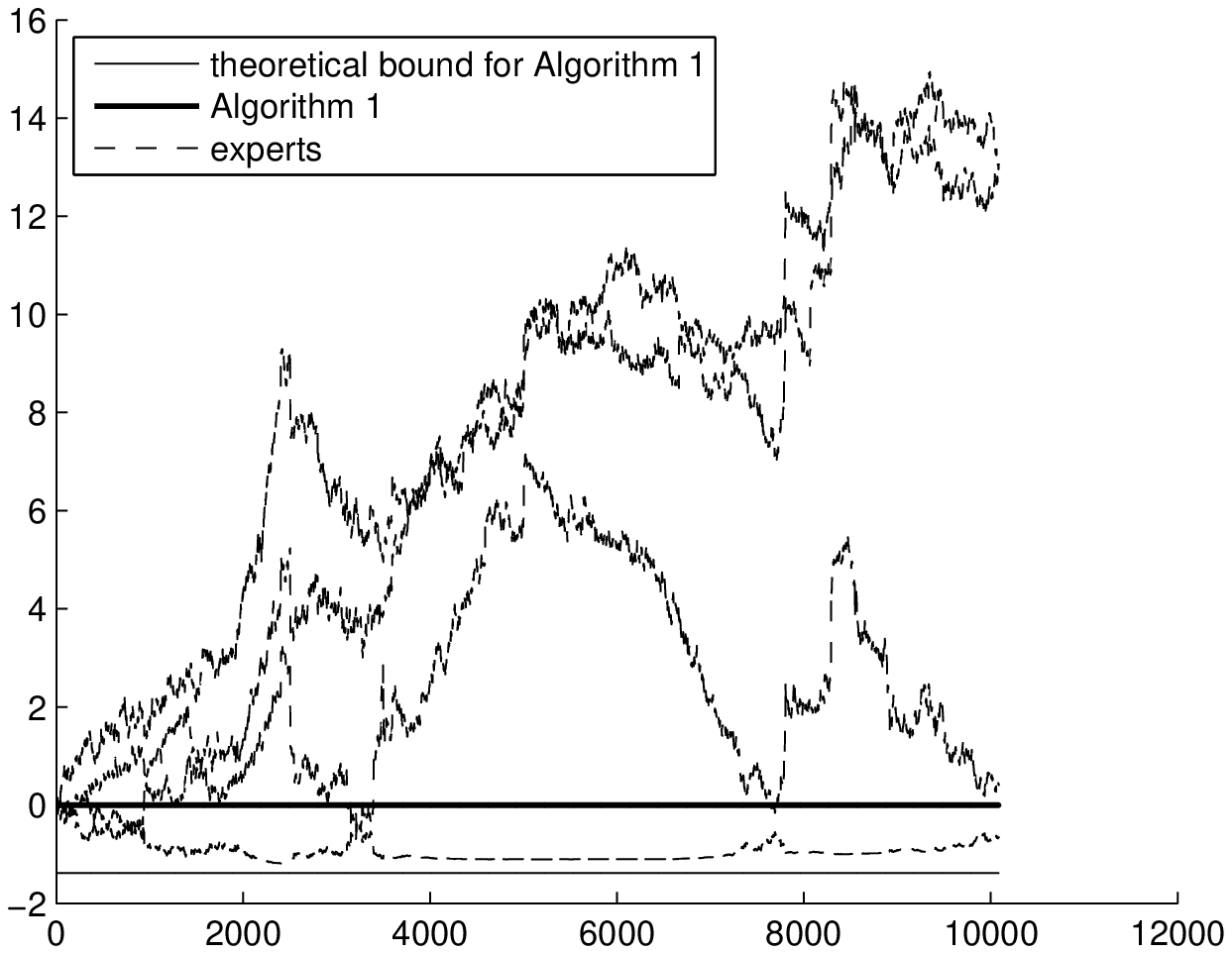}}
\caption{The difference between the cumulative loss of each of the $4$ bookmakers
  and of Algorithm~\ref{alg:SAA}
  on the tennis data.
  Now the theoretical bound is $-\ln 4$.}
\label{fig:tennis}
\end{center}
\end{figure}

Figure~\ref{fig:tennis} shows the results of another empirical study,
involving data about a large number of tennis tournaments in 2004, 2005, 2006, and 2007,
with the total number of matches 10,087.
The tournaments include, e.g.,  Australian Open, French Open, US Open, and Wimbledon;
the data is provided by Tennis-Data.
The matches are sorted by date, then by tournament, and then by the winner's name.
The data contain information about the winner of each match
and the betting odds of $4$ bookmakers for his/her win and for the opponent's win.
Therefore, now there are two possible observations
(player 1's win and player 2's win).
There are four bookmakers: Bet365, Centrebet, Expekt, and Pinnacle Sports.
The results in Figure~\ref{fig:tennis}
are presented in the same way as in Figure~\ref{fig:football}.

\begin{figure}[ht]
\begin{center}
\centerline{\includegraphics[width=\picturewidth]{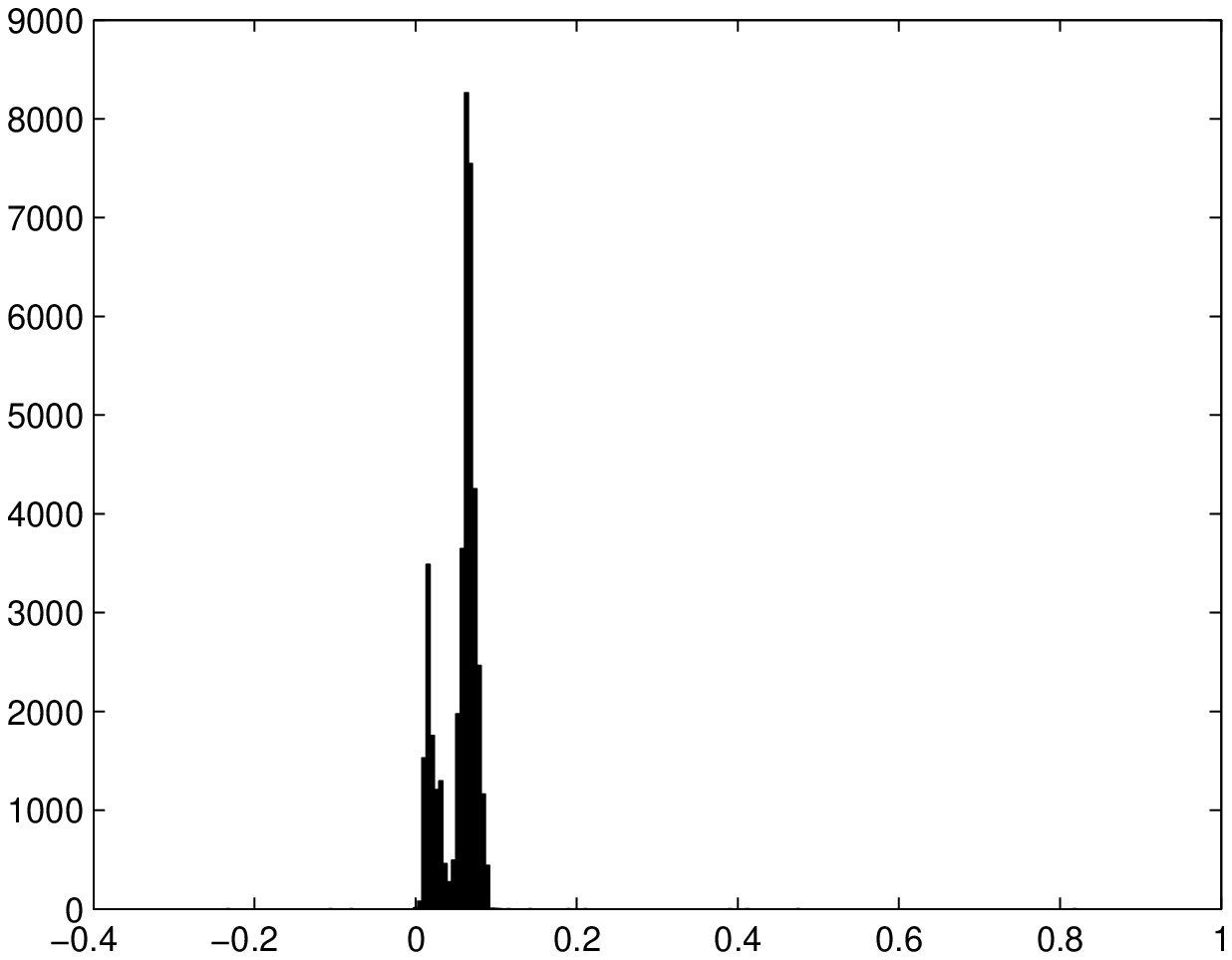}}
\caption{The overround distribution histogram for the tennis data.}
\label{fig:tennis-overround}
\end{center}
\end{figure}

Typical values of the overround are below $0.1$,
as shown in Figure \ref{fig:tennis-overround}
(analogous to Figure \ref{fig:football-overround}).


In both Figure~\ref{fig:football} and Figure~\ref{fig:tennis}
the cumulative loss of Algorithm~\ref{alg:SAA}
is close to the cumulative loss of the best expert,
despite the fact that some of the experts perform poorly.
The theoretical bound is not hopelessly loose for the football data
and is rather tight for the tennis data.
The pictures look exactly the same when Algorithm~\ref{alg:SAA}
is applied in the more realistic manner
where the experts' weights $w^k$ are not updated over the matches
that are played simultaneously.

Our second empirical study (Figure~\ref{fig:tennis})
is about binary prediction,
and so the algorithm of \citet{vovk:1990}
could have also been used
(and would have given similar results).
We included it since we are not aware of any empirical studies
even for the binary case.

For comparison with several other popular prediction algorithms,
see Appendix~\ref{app:comparisons}.
The data used for producing all the figures and tables
in this section and in Appendix~\ref{app:comparisons}
can be downloaded from \texttt{http://vovk.net/ICML2008}.

\section{Proof of Theorem \ref{thm:main}}
\label{sec:proof}

This proof will use some basic notions of elementary differential geometry,
especially those connected with the Gauss--Kronecker curvature of surfaces.
(The use of curvature in this kind of results is standard:
see, e.g., \citealt{vovk:1990} and \citealt{haussler/etal:1998}.)
All definitions that we will need can be found in, e.g., \citealt{thorpe:1979}.

A vector $f\in\bbbr^{\Omega}$
(understood to be a function $f:\Omega\to\bbbr$)
is a \emph{superprediction}
if there is $\gamma\in\Gamma$ such that, for all $\omega\in\Omega$,
$\lambda(\omega,\gamma)\le f(\omega)$;
the set $\Sigma$ of all superpredictions is the \emph{superprediction set}.
For each \emph{learning rate} $\eta>0$,
let $\Phi_{\eta}:\bbbr^{\Omega}\to(0,\infty)^{\Omega}$
be the homeomorphism defined by
\begin{equation}\label{eq:Phi}
  \Phi_{\eta}(f):\omega\in\Omega\mapsto e^{-\eta f(\omega)},
  \quad
  f\in\bbbr^{\Omega}.
\end{equation}
The image $\Phi_{\eta}(\Sigma)$ of the superprediction set
will be called the \emph{$\eta$-exponential superprediction set}.
It is known
that
\begin{equation*}
  L_N
  \le
  \min_{k=1,\ldots,K}
  L_N^k
  +
  \frac{\ln K}{\eta},
  \quad
  N=1,2,\ldots,
\end{equation*}
can be guaranteed
if and only if the $\eta$-exponential superprediction set is convex
(part ``if'' for all $K$ and part ``only if'' for $K\to\infty$
are proved in \citealt{vovk:1998game};
part ``only if'' for all $K$ is proved by Chris Watkins,
and the details can be found in Appendix~\ref{app:Watkins}).
Comparing this with (\ref{eq:can}) and (\ref{eq:cannot})
we can see that we are required to prove that
\begin{itemize}
\item
  $\Phi_{\eta}(\Sigma)$ is convex when $\eta\le1$;
\item
  $\Phi_{\eta}(\Sigma)$ is not convex when $\eta>1$.
\end{itemize}

Define the \emph{$\eta$-exponential superprediction surface}
to be the part of the boundary of the $\eta$-exponential superprediction set $\Phi_{\eta}(\Sigma)$
lying inside $(0,\infty)^{\Omega}$.
The idea of the proof is to check that, for all $\eta<1$,
the Gauss--Kronecker curvature of this surface is nowhere vanishing.
Even when this is done, however,
there is still uncertainty as to in which direction the surface is bulging
(towards the origin or away from it).
The standard argument
(as in \citealt{thorpe:1979}, Chapter 12, Theorem 6)
based on the continuity of the smallest principal curvature
shows that the $\eta$-exponential superprediction set
is bulging away from the origin for small enough $\eta$:
indeed, since it is true at some point,
it is true everywhere on the surface.
By the continuity in $\eta$ this is also true for all $\eta<1$.
Now, since
the $\eta$-exponential superprediction set is convex
for all $\eta<1$, it is also convex for $\eta=1$.

Let us now check
that the Gauss--Kronecker curvature of the $\eta$-exponential superprediction surface
is always positive when $\eta<1$
and is sometimes negative when $\eta>1$
(the rest of the proof, an elaboration of the above argument, will be easy).
Set $n:=\lvert\Omega\rvert$;
without loss of generality we assume $\Omega=\{1,\ldots,n\}$.

A convenient parametric representation of the $\eta$-exponential superprediction surface is
\begin{equation}\label{eq:surface}
  \begin{pmatrix}
    x^1\\x^2\\\vdots\\x^{n-1}\\x^n
  \end{pmatrix}
  =
  \begin{pmatrix}
    e^{-\eta((u^1-1)^2+(u^2)^2+\cdots+(u^n)^2)}\\
    e^{-\eta((u^1)^2+(u^2-1)^2+\cdots+(u^n)^2)}\\
    \vdots\\
    e^{-\eta((u^1)^2+\cdots+(u^{n-1}-1)^2+(u^n)^2)}\\
    e^{-\eta((u^1)^2+\cdots+(u^{n-1})^2+(u^n-1)^2)}
  \end{pmatrix},
\end{equation}
where $u^1,\ldots,u^{n-1}$ are the coordinates on the surface,
$u^1,\ldots,u^{n-1}\in(0,1)$ subject to $u^1+\cdots u^{n-1}<1$,
and $u^n$ is a shorthand for $1-u^1-\cdots-u^{n-1}$.
The derivative of (\ref{eq:surface}) in $u^1$ is
\begin{multline*}
  \frac{\partial}{\partial u^1}
  \begin{pmatrix}
    x^1\\x^2\\\vdots\\x^{n-1}\\x^n
  \end{pmatrix}
  =
  2\eta
  \begin{pmatrix}
    (u^n-u^1+1)
    e^{-\eta((u^1-1)^2+(u^2)^2+\cdots+(u^{n-1})^2+(u^n)^2)}\\
    (u^n-u^1)
    e^{-\eta((u^1)^2+(u^2-1)^2+\cdots+(u^{n-1})^2+(u^n)^2)}\\
    \vdots\\
    (u^n-u^1)
    e^{-\eta((u^1)^2+(u^2)^2+\cdots+(u^{n-1}-1)^2+(u^n)^2)}\\
    (u^n-u^1-1)
    e^{-\eta((u^1)^2+(u^2)^2+\cdots+(u^{n-1})^2+(u^n-1)^2)}
  \end{pmatrix}\\
  \propto
  \begin{pmatrix}
    (u^n-u^1+1) e^{2\eta u^1}\\
    (u^n-u^1) e^{2\eta u^2}\\
    \vdots\\
    (u^n-u^1) e^{2\eta u^{n-1}}\\
    (u^n-u^1-1) e^{2\eta u^n}
  \end{pmatrix},
\end{multline*}
the derivative in $u^2$ is
\begin{equation*}
  \frac{\partial}{\partial u^2}
  \begin{pmatrix}
    x^1\\x^2\\\vdots\\x^{n-1}\\x^n
  \end{pmatrix}
  \propto
  \begin{pmatrix}
    (u^n-u^2) e^{2\eta u^1}\\
    (u^n-u^2+1) e^{2\eta u^2}\\
    \vdots\\
    (u^n-u^2) e^{2\eta u^{n-1}}\\
    (u^n-u^2-1) e^{2\eta u^n}
  \end{pmatrix},
\end{equation*}
and so on,
up to
\begin{equation*}
  \frac{\partial}{\partial u^{n-1}}
  \begin{pmatrix}
    x^1\\x^2\\\vdots\\x^{n-1}\\x^n
  \end{pmatrix}
  \propto
  \begin{pmatrix}
    (u^n-u^{n-1}) e^{2\eta u^1}\\
    (u^n-u^{n-1}) e^{2\eta u^2}\\
    \vdots\\
    (u^n-u^{n-1}+1) e^{2\eta u^{n-1}}\\
    (u^n-u^{n-1}-1) e^{2\eta u^n}
  \end{pmatrix},
\end{equation*}
all coefficients of proportionality being equal and positive.

A normal vector to the surface can be found as
\begin{multline*}
  Z
  :=\\
  \begin{vmatrix}
    e_1 & \cdots & e_{n-1} & e_n\\
    (u^n-u^1+1) e^{2\eta u^1} & \cdots
      & (u^n-u^1) e^{2\eta u^{n-1}} & (u^n-u^1-1) e^{2\eta u^n}\\
    \vdots & \ddots & \vdots & \vdots\\
    (u^n-u^{n-1}) e^{2\eta u^1} & \cdots
      & (u^n-u^{n-1}+1) e^{2\eta u^{n-1}} & (u^n-u^{n-1}-1) e^{2\eta u^n}
  \end{vmatrix},
\end{multline*}
where $e_i$ is the $i$th vector in the standard basis of $\bbbr^n$.
The coefficient in front of $e_1$ is the $(n-1)\times(n-1)$ determinant
\begin{multline}\label{eq:coef-e1}
  \begin{vmatrix}
    (u^n-u^1) e^{2\eta u^2} & \cdots
      & (u^n-u^1) e^{2\eta u^{n-1}} & (u^n-u^1-1) e^{2\eta u^n}\\
    (u^n-u^2+1) e^{2\eta u^2} & \cdots
      & (u^n-u^2) e^{2\eta u^{n-1}} & (u^n-u^2-1) e^{2\eta u^n}\\
    \vdots & \ddots & \vdots & \vdots\\
    (u^n-u^{n-1}) e^{2\eta u^2} & \cdots
      & (u^n-u^{n-1}+1) e^{2\eta u^{n-1}} & (u^n-u^{n-1}-1) e^{2\eta u^n}
  \end{vmatrix}\\
  \propto
  e^{-2\eta u^1}
  \begin{vmatrix}
    u^n-u^1 & \cdots & u^n-u^1 & u^n-u^1-1\\
    u^n-u^2+1 & \cdots & u^n-u^2 & u^n-u^2-1\\
    \vdots & \ddots & \vdots & \vdots\\
    u^n-u^{n-1} & \cdots & u^n-u^{n-1}+1 & u^n-u^{n-1}-1
  \end{vmatrix}\\
  =
  e^{-2\eta u^1}
  \begin{vmatrix}
    1 & 1 & \cdots & 1 & u^n-u^1-1\\
    2 & 1 & \cdots & 1 & u^n-u^2-1\\
    1 & 2 & \cdots & 1 & u^n-u^3-1\\
    \vdots & \vdots & \ddots & \vdots & \vdots\\
    1 & 1 & \cdots & 2 & u^n-u^{n-1}-1
  \end{vmatrix}\\
  =
  e^{-2\eta u^1}
  \begin{vmatrix}
    1 & 1 & \cdots & 1 & u^n-u^1-1\\
    1 & 0 & \cdots & 0 & u^1-u^2\\
    0 & 1 & \cdots & 0 & u^1-u^3\\
    \vdots & \vdots & \ddots & \vdots & \vdots\\
    0 & 0 & \cdots & 1 & u^1-u^{n-1}
  \end{vmatrix}\\
  =
  e^{-2\eta u^1}
  \bigl(
    (-1)^n
    (u^n-u^1-1)
    +
    (-1)^{n+1}
    (u^1-u^2)\\
    +
    (-1)^{n+1}
    (u^1-u^3)
    +\cdots+
    (-1)^{n+1}
    (u^1-u^{n-1})
  \bigr)\\
  =
  e^{-2\eta u^1}
  (-1)^n
  \left(
    (u^2+u^3+\cdots+u^n)
    -
    (n-1) u^1
    -
    1
  \right)\\
  =
  -
  e^{-2\eta u^1}
  (-1)^nn u^1
  \propto
  u^1 e^{-2\eta u^1}
\end{multline}
(with a positive coefficient of proportionality,
$e^{2\eta}$, in the first $\propto$;
the third equality follows from the expansion of the determinant
along the last column and then along the first row).

Similarly, the coefficient in front of $e_i$ is proportional
(with the same coefficient of proportionality)
to $u^i e^{-2\eta u^i}$ for $i=2,\ldots,n-1$;
indeed,
the $(n-1)\times(n-1)$ determinant
representing the coefficient in front of $e_i$
can be reduced to the form analogous to (\ref{eq:coef-e1})
by moving the $i$th row to the top.

The coefficient in front of $e_n$ is proportional to
\begin{multline*}
  e^{-2\eta u^n}
  \begin{vmatrix}
    u^n-u^1+1 & u^n-u^1 & \cdots & u^n-u^1 & u^n-u^1\\
    u^n-u^2 & u^n-u^2+1 & \cdots & u^n-u^2 & u^n-u^2\\
    \vdots & \vdots & \ddots  & \vdots\\
    u^n-u^{n-2} & u^n-u^{n-2} & \cdots & u^n-u^{n-2}+1 & u^n-u^{n-2}\\
    u^n-u^{n-1} & u^n-u^{n-1} & \cdots & u^n-u^{n-1} & u^n-u^{n-1}+1
  \end{vmatrix}\\
  =
  e^{-2\eta u^n}
  \begin{vmatrix}
    1 & 0 & \cdots & 0 & u^n-u^1\\
    0 & 1 & \cdots & 0 & u^n-u^2\\
    \vdots & \vdots & \ddots & \vdots & \vdots\\
    0 & 0 & \cdots & 1 & u^n-u^{n-2}\\
    -1 & -1 & \cdots & -1 & u^n-u^{n-1}+1
  \end{vmatrix}\\
  =
  e^{-2\eta u^n}
  \begin{vmatrix}
    1 & 0 & \cdots & 0 & u^n-u^1\\
    0 & 1 & \cdots & 0 & u^n-u^2\\
    \vdots & \vdots & \ddots & \vdots & \vdots\\
    0 & 0 & \cdots & 1 & u^n-u^{n-2}\\
    0 & 0 & \cdots & 0 & n u^n
  \end{vmatrix}
  =
  n u^n e^{-2\eta u^n}
\end{multline*}
(with the coefficient of proportionality $e^{2\eta}(-1)^{n-1}$).

The Gauss--Kronecker curvature at the point with coordinates $(u^1,\ldots,u^{n-1})$
is proportional
(with a positive coefficient of proportionality,
possibly depending on the point)
to
\begin{equation}\label{eq:det}
  \begin{vmatrix}
    \frac{\partial Z\T}{\partial u^1}\\
    \vdots\\
    \frac{\partial Z\T}{\partial u^{n-1}}\\
    Z\T
  \end{vmatrix}
\end{equation}
(\citealt{thorpe:1979}, Chapter 12, Theorem 5,
with ${}^{\text{T}}$ standing for transposition).

A straightforward calculation allows us to rewrite determinant (\ref{eq:det})
(ignoring the positive coefficient $((-1)^{n-1} n e^{2\eta})^n$)
as
\begin{multline}\label{eq:final}
  \begin{vmatrix}
    (1-2\eta u^1) e^{-2\eta u^1} & 0 & \cdots & 0 & (2\eta u^n-1) e^{-2\eta u^n}\\
    0 & (1-2\eta u^2) e^{-2\eta u^2} & \cdots & 0 & (2\eta u^n-1) e^{-2\eta u^n}\\
    \vdots & \vdots & \ddots & \vdots & \vdots\\
    0 & 0 & \cdots & (1-2\eta u^{n-1}) e^{-2\eta u^{n-1}} & (2\eta u^n-1) e^{-2\eta u^n}\\
    u^1 e^{-2\eta u^1} & u^2 e^{-2\eta u^2} & \cdots & u^{n-1} e^{-2\eta u^{n-1}} & u^n e^{-2\eta u^n}
  \end{vmatrix}\\
  \propto
  \begin{vmatrix}
    1 - 2\eta u^1 & 0 & \cdots & 0 & 2\eta u^n - 1\\
    0 & 1 - 2\eta u^2 & \cdots & 0 & 2\eta u^n - 1\\
    \vdots & \ddots & \vdots & \vdots\\
    0 & 0 & \cdots & 1 - 2\eta u^{n-1} & 2\eta u^n - 1\\
    u^1 & u^2 & \cdots & u^{n-1} & u^n
  \end{vmatrix}\\
  =
  u^1 (1 - 2\eta u^2) (1 - 2\eta u^3) \cdots (1 - 2\eta u^n)\\
  + u^2 (1 - 2\eta u^1) (1 - 2\eta u^3) \cdots (1 - 2\eta u^n)
  + \cdots \\
  + u^n (1 - 2\eta u^1) (1 - 2\eta u^2) \cdots (1 - 2\eta u^{n-1})
\end{multline}
(with a positive coefficient of proportionality;
to avoid calculation of the parities of various permutations,
the reader might prefer to prove the last equality
by induction in $n$,
expanding the last determinant along the first column).
Our next goal is to show that the last expression in (\ref{eq:final})
is positive when $\eta<1$
but can be negative when $\eta>1$.

If $\eta>1$,
set $u^1=u^2:=1/2$ and $u^3=\cdots=u^n:=0$.
The last expression in (\ref{eq:final}) becomes negative.
It will remain negative if $u^1$ and $u^2$ are sufficiently close to $1/2$
and $u^3,\ldots,u^n$ are sufficiently close to $0$.

It remains to consider the case $\eta<1$.
Set $t_i:=1-2\eta u^i$, $i=1,\ldots,n$;
the constraints on the $t_i$ are
\begin{equation}\label{eq:constraints}
  \begin{aligned}
    -1 < 1-2\eta < t_i &< 1,
    \quad
    i=1,\ldots,n,\\
    t_1+\cdots+t_n &= n-2\eta > n-2.
  \end{aligned}
\end{equation}
Our goal is to prove
\begin{equation*}
  (1-t_1) t_2 t_3 \cdots t_n
  + \cdots +
  (1-t_n) t_1 t_2 \cdots t_{n-1}
  >
  0,
\end{equation*}
i.e.,
\begin{equation}\label{eq:1}
  t_2 t_3 \cdots t_n
  + \cdots +
  t_1 t_2 \cdots t_{n-1}
  >
  n t_1 \cdots t_n.
\end{equation}
This reduces to
\begin{equation}\label{eq:2}
  \frac{1}{t_1}
  + \cdots +
  \frac{1}{t_n}
  >
  n
\end{equation}
if $t_1\cdots t_n>0$,
and to
\begin{equation}\label{eq:3}
  \frac{1}{t_1}
  + \cdots +
  \frac{1}{t_n}
  <
  n
\end{equation}
if $t_1\cdots t_n<0$.
The remaining case is where some of the $t_i$ are zero;
for concreteness, let $t_n=0$.
By (\ref{eq:constraints}) we have $t_1+\cdots+t_{n-1}>n-2$,
and so all of $t_1,\ldots,t_{n-1}$ are positive;
this shows that (\ref{eq:1}) is indeed true.

Let us prove (\ref{eq:2}).
Since $t_1\cdots t_n>0$, all of $t_1,\ldots,t_n$ are positive
(if two of them were negative,
the sum $t_1+\cdots+t_n$ would be less than $n-2$;
cf.\ (\ref{eq:constraints})).
Therefore,
\begin{equation*}
  \frac{1}{t_1}
  + \cdots +
  \frac{1}{t_n}
  >
  \underbrace{1+\cdots+1}_{\text{$n$ times}}
  =
  n.
\end{equation*}

To establish (\ref{eq:1})
it remains to prove (\ref{eq:3}).
Suppose, without loss of generality,
that $t_1>0$, $t_2>0$,\ldots, $t_{n-1}>0$, and $t_n<0$.
We will prove a slightly stronger statement
allowing $t_1,\ldots,t_{n-2}$ to take value $1$
and removing the lower bound on $t_n$.
Since the function $t\in(0,1]\mapsto1/t$ is convex,
we can also assume, without loss of generality,
$t_1=\cdots=t_{n-2}=1$.
Then $t_{n-1}+t_n>0$,
and so
\begin{equation*}
  \frac{1}{t_{n-1}}
  +
  \frac{1}{t_n}
  <
  0;
\end{equation*}
therefore,
\begin{equation*}
  \frac{1}{t_1}
  + \cdots +
  \frac{1}{t_{n-2}}
  +
  \frac{1}{t_{n-1}}
  +
  \frac{1}{t_n}
  <
  n-2
  <
  n.
\end{equation*}

Finally, let us check that the positivity of the Gauss--Kronecker curvature
implies the convexity of the $\eta$-exponential superprediction set
in the case $\eta\le1$,
and the lack of positivity of the Gauss--Kronecker curvature
implies the lack of convexity of the $\eta$-exponential superprediction set
in the case $\eta>1$.
The $\eta$-exponential superprediction surface
will be oriented by choosing the normal vector field
directed towards the origin.
This can be done since
\begin{equation}\label{eq:two}
  \begin{pmatrix}
    x^1\\\vdots\\x^n
  \end{pmatrix}
  \propto
  \begin{pmatrix}
    e^{2\eta u^1}\\
    \vdots\\
    e^{2\eta u^n}
  \end{pmatrix},
  \quad
  Z
  \propto
  (-1)^{n-1}
  \begin{pmatrix}
    u^1 e^{-2\eta u^1}\\
    \vdots\\
    u^n e^{-2\eta u^n}
  \end{pmatrix},
\end{equation}
with both coefficients of proportionality positive
(cf.\ (\ref{eq:surface}) and the bottom row of the first determinant in (\ref{eq:final})),
and the sign of the scalar product of the two vectors on the right-hand sides in (\ref{eq:two})
does not depend on the point $(u^1,\ldots,u^{n-1})$.
Namely, we take $(-1)^n Z$ as the normal vector field directed towards the origin.
The Gauss--Kronecker curvature will not change sign after the re-orientation:
if $n$ is even, the new orientation coincides with the old,
and for odd $n$ the Gauss--Kronecker curvature does not depend on the orientation.

In the case $\eta>1$,
the Gauss--Kronecker curvature is negative at some point,
and so the $\eta$-exponential superprediction set is not convex
(\citealt{thorpe:1979}, Chapter 13, Theorem 1 and its proof).

It remains to consider the case $\eta\le1$.
Because of the continuity of the $\eta$-exponential superprediction surface in $\eta$
we can and will assume, without loss of generality, that $\eta<1$.

Let us first check that the smallest principal curvature
$$k_1=k_1(u^1,\ldots,u^{n-1},\eta)$$
of the $\eta$-exponential superprediction surface
is always positive
(among the arguments of $k_1$ we list not only the coordinates $u^1,\ldots,u^{n-1}$
of a point on the surface (\ref{eq:surface})
but also the learning rate $\eta\in(0,1)$).
At least at some $(u^1,\ldots,u^{n-1},\eta)$
the value of $k_1(u^1,\ldots,u^{n-1},\eta)$ is positive:
take a sufficiently small $\eta$
and the point on the surface (\ref{eq:surface})
at which the maximum of $x^1+\cdots+x^n$ is attained
(the point of the $\eta$-exponential superprediction set
at which the maximum is attained will lie on the surface
since the maximum is attained
at $(x^1,\ldots,x^n)=(1,\ldots,1)$ when $\eta=0$).
Therefore,
for all $(u^1,\ldots,u^{n-1},\eta)$
the value of $k_1(u^1,\ldots,u^{n-1},\eta)$ is positive:
if $k_1$ had different signs
at two points in the set
\begin{multline}\label{eq:set}
  \bigl\{
    (u^1,\ldots,u^{n-1},\eta)
    \st
    u^1\in(0,1),\ldots,u^{n-1}\in(0,1),\\
    u^1+\cdots+u^{n-1}<1,
    \eta\in(0,1)
  \bigr\},
\end{multline}
we could connect these points by a continuous curve
lying completely inside (\ref{eq:set});
at some point on the curve, $k_1$ would be zero,
in contradiction to the positivity of the Gauss--Kronecker curvature
$k_1\cdots k_{n-1}$.

Now it is easy to show that the $\eta$-exponential superprediction set is convex.
Suppose there are two points $A$ and $B$
on the $\eta$-exponential superprediction surface
such that the interval $[A,B]$ contains points outside the $\eta$-exponential superprediction set.
The intersection of the plane $OAB$,
where $O$ is the origin,
with the $\eta$-exponential superprediction surface
is a planar curve;
the curvature of this curve at some point between $A$ and $B$
will be negative
(remember that the curve is oriented
by directing the normal vector field towards the origin),
contradicting the positivity of $k_1$ at that point.

\section{Derivation of the prediction algorithm}
\label{sec:algorithm}

To achieve the loss bound (\ref{eq:can}) in Theorem \ref{thm:main}
Learner can use, as discussed earlier, the strong aggregating algorithm
(see, e.g., \citealt{vovk:2001competitive}, Section 2.1, (15))
with $\eta=1$.
In this section we will find a substitution function
for the strong aggregating algorithm for the Brier game with $\eta\le1$,
which is the only component of the algorithm not described explicitly
in \citet{vovk:2001competitive}.
Our substitution function will not require
that its input, the generalized prediction,
should be computed from the normalized distribution $(w^k)_{k=1}^K$ on the experts;
this is a valuable feature for generalizations
to an infinite number of experts
(as demonstrated in, e.g., \citealt{vovk:2001competitive}, Appendix A.1).

Suppose that we are given a generalized prediction
$(l_1,\ldots,l_n)\T$ computed by the aggregating pseudo-algorithm
from a normalized distribution on the experts.
Since $(l_1,\ldots,l_n)\T$ is a superprediction
(remember that we are assuming $\eta\le1$),
we are only required to find a permitted prediction
\begin{equation}\label{eq:lambda}
  \begin{pmatrix}
    \lambda_1\\\lambda_2\\\vdots\\\lambda_n
  \end{pmatrix}
  =
  \begin{pmatrix}
    (u^1-1)^2+(u^2)^2+\cdots+(u^n)^2\\
    (u^1)^2+(u^2-1)^2+\cdots+(u^n)^2\\
    \vdots\\
    (u^1)^2+(u^2)^2+\cdots+(u^n-1)^2
  \end{pmatrix}
\end{equation}
(cf.\ (\ref{eq:surface}))
satisfying
\begin{equation}\label{eq:required}
  \lambda_1\le l_1,
  \ldots,
  \lambda_n\le l_n.
\end{equation}

Now suppose we are given a generalized prediction
$(L_1,\ldots,L_n)\T$ computed by the aggregating pseudo-algorithm
from an unnormalized distribution on the experts;
in other words,
we are given
\begin{equation*}
  \begin{pmatrix}
    L_1\\\vdots\\L_n
  \end{pmatrix}
  =
  \begin{pmatrix}
    l_1+c\\\vdots\\l_n+c
  \end{pmatrix}
\end{equation*}
for some $c\in\bbbr$.
To find (\ref{eq:lambda}) satisfying (\ref{eq:required})
we can first find the largest $t\in\bbbr$
such that $(L_1-t,\ldots,L_n-t)\T$ is still a superprediction
and then find (\ref{eq:lambda}) satisfying
\begin{equation}\label{eq:old-constraints}
  \lambda_1\le L_1-t,
  \ldots,
  \lambda_n\le L_n-t.
\end{equation}
Since $t\ge c$,
it is clear that $(\lambda_1,\ldots,\lambda_n)\T$
will also satisfy the required (\ref{eq:required}).

\begin{proposition}\label{prop:main}
  Define $s\in\bbbr$ by the requirement
  \begin{equation}\label{eq:requirement}
    \sum_{i=1}^n
    (s-L_i)^+
    =
    2.
  \end{equation}
  The unique solution to the optimization problem $t\to\max$
  under the constraints (\ref{eq:old-constraints})
  with $\lambda_1,\ldots,\lambda_n$ as in (\ref{eq:lambda})
  will be
  \begin{align}
    u^i
    &=
    \frac{(s-L_i)^+}{2},
    \enspace
    i=1,\ldots,n,\label{eq:u}\\
    t
    &=
    s-1-(u^1)^2-\cdots-(u^n)^2.
    \label{eq:t}
  \end{align}
\end{proposition}
There exists a unique $s$ satisfying (\ref{eq:requirement})
since the left-hand side of (\ref{eq:requirement})
is a continuous, increasing (strictly increasing when positive) and unbounded above
function of $s$.
The substitution function is given by (\ref{eq:u}).

\ifarXiv
  \begin{proof}[Proof of Proposition \ref{prop:main}]
\fi
\ifJOURNAL
  \begin{proof}{Proof of Proposition \ref{prop:main}}
\fi
  Let us denote the $u^i$ and $t$ defined by (\ref{eq:u}) and (\ref{eq:t})
  as $\overline{u}^i$ and $\overline{t}$, respectively.
  To see that they satisfy the constraints (\ref{eq:old-constraints}),
  notice that the $i$th constraint can be spelt out as
  $$
    (\overline{u}^1)^2+\cdots+(\overline{u}^n)^2
    -
    2\overline{u}^i
    +
    1
    \le
    L_i - \overline{t},
  $$
  which immediately follows from (\ref{eq:u}) and (\ref{eq:t}).
  As a by-product, we can see that the inequality becomes an equality,
  i.e.,
  \begin{equation}\label{eq:t-overline}
    \overline{t}
    =
    L_i - 1 + 2\overline{u}^i
    -
    (\overline{u}^1)^2-\cdots-(\overline{u}^n)^2,
  \end{equation}
  for all $i$ with $\overline{u}^i>0$.

  We can rewrite (\ref{eq:old-constraints}) as
  \begin{equation}\label{eq:new-constraints}
    \left\{
      \begin{matrix}
        t \le L_1 - 1 + 2u^1 - (u^1)^2 - \cdots - (u^n)^2,\\
        \vdots\\
        t \le L_n - 1 + 2u^n - (u^1)^2 - \cdots - (u^n)^2,
      \end{matrix}
    \right.
  \end{equation}
  and our goal is to prove that these inequalities imply $t<\overline{t}$
  (unless $u^1=\overline{u}^1,\ldots,u^n=\overline{u}^n$).
  Choose $\overline{u}^i$
  (necessarily $\overline{u}^i>0$
  unless $u^1=\overline{u}^1,\ldots,u^n=\overline{u}^n$;
  in the latter case, however, we can, and will, also choose $\overline{u}^i>0$)
  for which $\epsilon_i:=\overline{u}^i-u^i$ is maximal.
  Then every value of $t$ satisfying (\ref{eq:new-constraints}) will also satisfy
  \begin{align*}
    t
    &\le
    L_i
    -
    1
    +
    2u^i
    -
    \sum_{j=1}^n
    (u^j)^2\\
    &=
    L_i
    -
    1
    +
    2\overline{u}^i
    -
    2\epsilon_i
    -
    \sum_{j=1}^n
    (\overline{u}^j)^2
    +
    2\sum_{j=1}^n
    \epsilon_j \overline{u}^j
    -
    \sum_{j=1}^n
    \epsilon_j^2\\
    &\le
    L_i
    -
    1
    +
    2\overline{u}^i
    -
    \sum_{j=1}^n
    (\overline{u}^j)^2
    -
    \sum_{j=1}^n
    \epsilon_j^2
    \le
    \overline{t},
  \end{align*}
  with the last $\le$ following from (\ref{eq:t-overline})
  and becoming $<$
  when not all $u^j$ coincide with $\overline{u}^j$.
\end{proof}

The detailed description of the resulting prediction algorithm
was given as Algorithm~\ref{alg:SAA} in Section~\ref{sec:main}.
As discussed, that algorithm uses the generalized prediction $G_N(\omega)$
computed from unnormalized weights.

\section{Conclusion}

In this paper we only considered the simplest prediction problem
for the Brier game:
competing with a finite pool of experts.
In the case of square-loss regression,
it is possible to find efficient closed-form prediction algorithms
competitive with linear functions
(see, e.g., \citealt{cesabianchi/lugosi:2006}, Chapter 11).
Such algorithms can often be ``kernelized''
to obtain prediction algorithms
competitive with reproducing kernel Hilbert spaces
of prediction rules.
This would be an appealing research programme
in the case of the Brier game as well.

\subsection*{Acknowledgments}

We are grateful to Football-Data and Tennis-Data
for providing access to the data used in this paper.
This work was partly supported by EPSRC
(grant EP/F002998/1).
Comments by Alexey Chernov, Yuri Kalnishkan, Alex Gammerman, Bob Vickers,
and the anonymous referees for the conference version
have helped us improve the presentation.
The latter also suggested comparing our results
to the Weighted Average Algorithm and the Hedge algorithm.

\appendix

\section{Watkins's theorem}
\label{app:Watkins}

Watkins's theorem is stated in \cite{vovk:1999derandomizing} (Theorem 8)
not in sufficient generality:
it presupposes that the loss function is perfectly mixable.
The proof, however, shows that this assumption is irrelevant
(it can be made part of the conclusion),
and the goal of this appendix is to give a self-contained statement
of a suitable version of the theorem.

In this appendix we will use a slightly more general notion
of a game of prediction $(\Omega,\Gamma,\lambda)$:
namely, the loss function $\lambda:\Omega\times\Gamma\to\overline{\bbbr}$
is now allowed to take values in the extended real line
$\overline{\bbbr}:=\bbbr\cup\{-\infty,\infty\}$
(although the value $-\infty$ will be later disallowed).

Partly following \cite{vovk:1998game},
for each $K=1,2,\ldots$ and each $a>0$
we consider the following perfect-information game $\GGG_K(a)$
(the ``global game'')
between two players, Learner and Environment.
Environment is a team of $K+1$ players called Expert~1 to Expert~$K$
and Reality,
who play with Learner according to Protocol \ref{prot:PEA}.
Learner wins if, for all $N=1,2,\ldots$ and all $k\in\{1,\ldots,K\}$,
\begin{equation}\label{eq:goal}
  L_N \le L_N^k + a;
\end{equation}
otherwise, Environment wins.
It is possible that $L_N=\infty$ or $L_N^k=\infty$ in (\ref{eq:goal});
the interpretation of inequalities involving infinities is natural.

For each $K$
we will be interested in the set of those $a>0$
for which Learner has a winning strategy
in the game $\GGG_K(a)$
(we will denote this by ${\rm L}\smile\GGG_K(a)$).
It is obvious that
\begin{equation*}
  {\rm L}\smile\GGG_K(a)
  \;\&\;
  a'>a
  \Longrightarrow
  {\rm L}\smile\GGG_K(a');
\end{equation*}
therefore, 
for each $K$ there exists a unique \emph{borderline value} $a_K$
such that ${\rm L}\smile\GGG_K(a)$ holds when $a>a_K$
and fails when $a<a_K$.
It is possible that $a_K=\infty$
(but remember that we are only interested in finite values of $a$).

These are our assumptions about the game of prediction
(similar to those in \cite{vovk:1998game}):
\begin{itemize}
\item
  $\Gamma$ is a compact topological space;
\item
  for each $\omega\in\Omega$,
  the function $\gamma\in\Gamma\mapsto\lambda(\omega,\gamma)$
  is continuous
  ($\overline{\bbbr}$ is equipped with the standard topology);
\item
  there exists $\gamma\in\Gamma$ such that,
  for all $\omega\in\Omega$,
  $\lambda(\omega,\gamma)<\infty$;
\item
  the function $\lambda$ is bounded below.
\end{itemize}

We say that the game of prediction $(\Omega,\Gamma,\lambda)$
is \emph{$\eta$-mixable}, where $\eta>0$, if
\begin{multline}\label{eq:mixture}
  \forall \gamma_1\in\Gamma,\gamma_2\in\Gamma,\alpha\in[0,1] \;
  \exists \delta\in\Gamma \;
  \forall \omega\in\Omega
  \colon\\
  e^{-\eta\lambda(\omega,\delta)}
  \ge
  \alpha
  e^{-\eta\lambda(\omega,\gamma_1)}
  +
  (1-\alpha)
  e^{-\eta\lambda(\omega,\gamma_2)}.
\end{multline}
In the case of finite $\Omega$,
this condition says that the image of the superprediction set
under the mapping $\Phi_{\eta}$ (see (\ref{eq:Phi}))
is convex.
The game of prediction is \emph{perfectly mixable}
if it is $\eta$-mixable for some $\eta>0$.

It follows from \cite{hardy/etal:1952}
(Theorem 92, applied to the means $\mathfrak{M}_{\phi}$ with $\phi(x)=e^{-\eta x}$)
that if the prediction game is $\eta$-mixable
it will remain $\eta'$-mixable for any positive $\eta'<\eta$.
(For another proof,
see the end of the proof of Lemma 9 in \cite{vovk:1998game}.)
Let $\eta^*$ be the supremum of the $\eta$
for which the prediction game is $\eta$-mixable
(with $\eta^*:=0$ when the game is not perfectly mixable).
The compactness of $\Gamma$ implies that the prediction game is $\eta^*$-mixable.

\begin{theorem}[Chris Watkins]\label{thm:Watkins}
  For any $K\in\{1,2,\ldots\}$,
  \begin{equation*}
    a_K
    =
    \frac{\ln K}{\eta^*}.
  \end{equation*}
  In particular,
  $a_K<\infty$ if and only if the game is perfectly mixable.
\end{theorem}
The theorem does not say explicitly, but it is easy to check, that ${\rm L}\smile\GGG_K(a_K)$:
this follows both from general considerations
(cf.\ Lemma 3 in \cite{vovk:1998game})
and from the fact that the SAA wins $\GGG_K(a_K)=\GGG_K(\ln K/\eta^*)$.
\ifarXiv
  \begin{proof}[Proof of Theorem \ref{thm:Watkins}]
\fi
\ifJOURNAL
  \begin{proof}{Proof of Theorem \ref{thm:Watkins}}
\fi
  The proof will use some notions and notation
  used in the statement and proof of Theorem 1 of \cite{vovk:1998game}.
  Without loss of generality
  we can, and will, assume that the loss function satisfies $\lambda>1$
  (add a suitable constant to $\lambda$ if needed).
  Therefore, Assumption 4 of \cite{vovk:1998game}
  (the only assumption in \cite{vovk:1998game} not directly made in this paper)
  is satisfied.
  In view of the fact that ${\rm L}\smile\GGG_K(\ln K/\eta^*)$,
  we only need to show that ${\rm L}\smile\GGG_K(a)$ does not hold for $a<\ln K/\eta^*$.
  Fix $a < \ln K/\eta^*$.

  The separation curve, as defined in \cite{vovk:1998game},
  consists of the points $(c(\beta),c(\beta)/\eta)\in[0,\infty)^2$,
  where $\beta:=e^{-\eta}$ and $\eta$ ranges over $[0,\infty]$
  (see \cite{vovk:1998game}, Theorem 1).
  Since the two-fold convex mixture in (\ref{eq:mixture})
  can be replaced by any finite convex mixture
  (apply two-fold mixtures repeatedly),
  setting $\eta:=\eta^*$ shows that the point $(1,1/\eta^*)$
  is Northeast of (actually belongs to) the separation curve.
  On the other hand, the point $(1,a/\ln K)$ is Southwest and outside of the separation curve
  (use Lemmas 8--12 of \cite{vovk:1998game}).
  Therefore,
  E ($=$Environment) has a winning strategy in the game $\GGG(1,a/\ln K)$,
  as defined in \cite{vovk:1998game}.
  It is easy to see from the proof of Theorem 1 in \cite{vovk:1998game}
  that the definition of the game $\GGG$ in \cite{vovk:1998game}
  can be modified, without changing the conclusion about $\GGG(1,a/\ln K)$,
  by replacing the line

  \medskip

  \noindent
  \hspace*{6mm}
  E chooses $n\ge1$ \{size of the pool\}

  \medskip

  \noindent
  in the protocol on p.~153 of \cite{vovk:1998game}
  by

  \medskip

  \noindent
  \hspace*{6mm}
  E chooses $n^*\ge1$ \{lower bound on the size of the pool\}\\
  \hspace*{6mm}
  L chooses $n\ge n^*$ \{size of the pool\}

  \medskip

  \noindent
  (indeed, the proof in Section~6 of \cite{vovk:1998game}
  only requires that there should be sufficiently many experts).
  Let $n^*$ be the first move by Environment
  according to her winning strategy.

  Now suppose ${\rm L}\smile\GGG_K(a)$.
  From the fact that there exists Learner's strategy ${\cal L}_1$
  winning $\GGG_K(a)$ we can deduce:
  there exists Learner's strategy ${\cal L}_2$
  winning $\GGG_{K^2}(2a)$
  (we can split the $K^2$ experts into $K$ groups of $K$,
  merge the experts' decisions in each group with ${\cal L}_1$,
  and finally merge the groups' decisions with ${\cal L}_1$);
  there exists Learner's strategy ${\cal L}_3$
  winning $\GGG_{K^3}(3a)$
  (we can split the $K^3$ experts into $K$ groups of $K^2$,
  merge the experts' decisions in each group
  with ${\cal L}_2$,
  and finally merge the groups' decisions with ${\cal L}_1$);
  and so on.
  When the number $K^m$ of experts exceeds $n^*$,
  we obtain a contradiction:
  Learner can guarantee
  \begin{equation*}
    L_N
    \le
    L_N^k
    +
    ma
  \end{equation*}
  for all $N$ and all $K^m$ experts $k$,
  and Environment can guarantee that
  \begin{equation*}
    L_N
    >
    L_N^k
    +
    \frac{a}{\ln K}
    \ln(K^m)
    =
    L_N^k
    +
    ma
  \end{equation*}
  for some $N$ and $k$.
\end{proof}

\section{Comparison with other prediction algorithms}
\label{app:comparisons}

Other popular algorithms for prediction with expert advice
that could be used instead of Algorithm \ref{alg:SAA}
in our empirical studies reported in Section \ref{sec:experimental}
are, among others,
Kivinen and Warmuth's \citeyear{kivinen/warmuth:1999}
Weighted Average Algorithm (WdAA),
Kalnishkan and Vyugin's \citeyear{kalnishkan/vyugin:2005}
Weak Aggregating Algorithm (WkAA),
and Freund and Schapire's \citeyear{freund/schapire:1997}
Hedge algorithm (HA).
In this appendix we consider these three algorithms
and three more naive algorithms
(which, nevertheless, perform surprisingly well).

The Weighted Average Algorithm is very similar
to the Strong Aggregating Algorithm (SAA) used in this paper:
the WdAA maintains the same weights for the experts as the SAA,
and the only difference is that the WdAA merges the experts' predictions
by averaging them according to their weights,
whereas the SAA uses a more complicated ``minimax optimal'' merging scheme
(given by (\ref{eq:u}) for the Brier game).
The performance guarantee for the WdAA applied to the Brier game
is weaker than the optimal (\ref{eq:can}),
but of course this does not mean that its empirical performance
is necessarily worse than that of the SAA
(i.e., Algorithm~\ref{alg:SAA}).
Figures \ref{fig:WdAAFootball-5} and \ref{fig:WdAATennis-4}
show the performance of this algorithm,
in the same format as before
(see Figures \ref{fig:football} and \ref{fig:tennis}).
We can see that for the football data
the maximal difference between the cumulative loss of the WdAA
and the cumulative loss of the best expert
is larger that for Algorithm \ref{alg:SAA}
but still well within the optimal bound $\ln K$ given by (\ref{eq:can}).
For the tennis data the maximal difference
is about twice as large as for Algorithm~\ref{alg:SAA},
violating the optimal bound $\ln K$.

\begin{figure}
\begin{center}
\centerline{\includegraphics[width=\picturewidth]{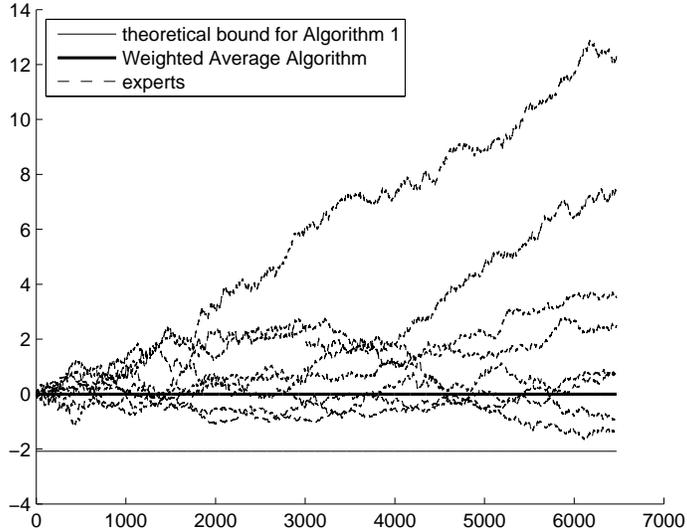}}
\caption{The difference between the cumulative loss of each of the $8$ bookmakers
  and of the Weighted Average Algorithm (WdAA) on the football data.
  The chosen value of the parameter $c=1/\eta$ for the WdAA,
  $c:=16/3$,
  minimizes its theoretical loss bound.
  The theoretical lower bound $-\ln 8\approx-2.0794$ for Algorithm~\ref{alg:SAA} is also shown
  (the theoretical lower bound for the Weighted Average Algorithm,
  $-11.0904$, can be extracted from Table \ref{tab:football-difference} below).}
\label{fig:WdAAFootball-5}
\end{center}
\end{figure}

\begin{figure}
\begin{center}
\centerline{\includegraphics[width=\picturewidth]{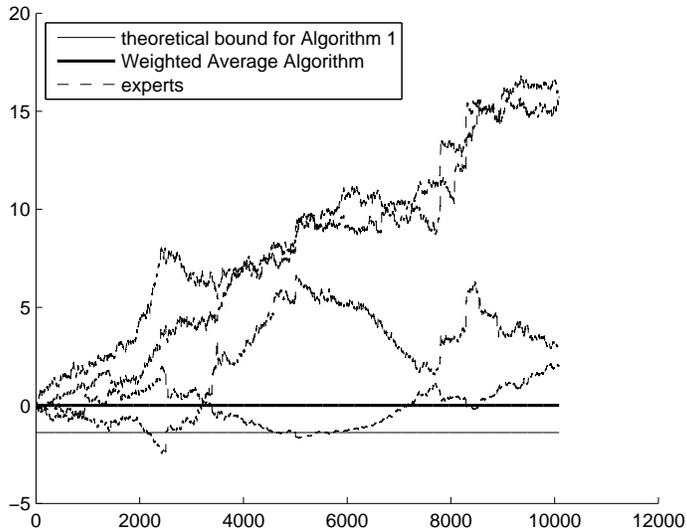}}
\caption{The difference between the cumulative loss of each of the $4$ bookmakers
  and of the WdAA for $c:=4$ on the tennis data.}
\label{fig:WdAATennis-4}
\end{center}
\end{figure}

In its most basic form
(\cite{kivinen/warmuth:1999}, the beginning of Section~6),
the WdAA works in the following protocol.
At each step each expert, Learner, and Reality
choose an element of the unit ball in $\bbbr^n$,
and the loss function is the squared distance
between the decision (Learner's or an expert's move)
and the observation (Reality's move).
This covers the Brier game with $\Omega=\{1,\ldots,n\}$,
each observation $\omega\in\Omega$ represented as the vector
$(\delta_{\omega}\{1\},\ldots,\delta_{\omega}\{n\})$,
and each decision $\gamma\in\PPP(\Omega)$ represented as the vector
$(\gamma\{1\},\ldots,\gamma\{n\})$.
However, in the Brier game the decision makers' moves
are known to belong to the simplex
$\{(u^1,\ldots,u^n)\in[0,\infty)^n\st\sum_{i=1}^n u^i = 1\}$,
and Reality's move is known to be one of the vertices of this simplex.
Therefore, we can optimize the ball radius
by considering the smallest ball containing the simplex
rather than the unit ball.
This is what we did for the results reported here
(although the results reported in the conference version of this paper
\cite{vovk/zhdanov:2008ICML}
are for the WdAA applied to the unit cube in $\bbbr^n$).
The radius of the smallest ball is
$$
  R
  :=
  \sqrt{1-\frac{1}{n}}
  \approx
  \begin{cases}
    0.8165 & \text{if $n=3$}\\
    0.7071 & \text{if $n=2$}\\
    1 & \text{if $n$ is large}.
  \end{cases}
$$
As described in \cite{kivinen/warmuth:1999},
the WdAA is parameterized by $c:=1/\eta$ instead of $\eta$,
and the optimal value of $c$ is $c=8R^2$,
leading to the guaranteed loss bound
\begin{equation*}
  L_N
  \le
  \min_{k=1,\ldots,K}
  L_N^k
  +
  8R^2
  \ln K
\end{equation*}
for all $N=1,2,\ldots$
(see \cite{kivinen/warmuth:1999}, Section~6).
This is significantly looser than the bound (\ref{eq:can})
for Algorithm~\ref{alg:SAA}.

The values $c=16/3$ and $c=4$
used in Figures~\ref{fig:WdAAFootball-5} and~\ref{fig:WdAATennis-4},
respectively,
are obtained by minimizing the WdAA's performance guarantee,
but minimizing a loose bound might not be such a good idea.
Figure~\ref{fig:WdAAFootball-all} shows the maximal difference
\begin{equation}\label{eq:maximal-football}
  \max_{N=1,\ldots,6473}
  \left(
    L_N(c)
    -
    \min_{k=1,\ldots,8}
    L_N^k
  \right),
\end{equation}
where $L_N(c)$ is the loss of the WdAA with parameter $c$
on the football data over the first $N$ steps
and $L_N^k$ is the analogous loss of the $k$th expert,
as a function of $c$.
Similarly, Figure~\ref{fig:WdAATennis-all} shows the maximal difference
\begin{equation}\label{eq:maximal-tennis}
  \max_{N=1,\ldots,10087}
  \left(
    L_N(c)
    -
    \min_{k=1,\ldots,4}
    L_N^k
  \right)
\end{equation}
for the tennis data.
And indeed, in both cases the value of $c$ minimizing the empirical loss
is far from the value minimizing the bound;
as could be expected,
the empirical optimal value for the WdAA is not so different
from the optimal value for Algorithm \ref{alg:SAA}.
The following two figures,
\ref{fig:AAFootball-all} and \ref{fig:AATennis-all},
demonstrate that there is no such anomaly for Algorithm \ref{alg:SAA}.

\begin{figure}
\begin{center}
\centerline{\includegraphics[width=\picturewidth]{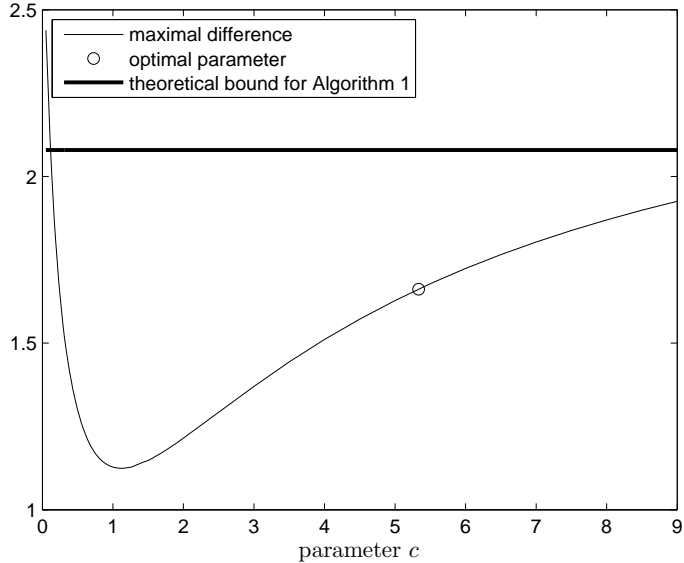}}
\caption{The maximal difference (\ref{eq:maximal-football})
  for the WdAA as function of the parameter $c$ on the football data.
  The theoretical guarantee $\ln 8$ for the maximal difference
  for Algorithm~\ref{alg:SAA} is also shown
  (the theoretical guarantee for the WdAA,
  $11.0904$, is given in Table \ref{tab:football-difference}).}
\label{fig:WdAAFootball-all}
\end{center}
\end{figure}

\begin{figure}
\begin{center}
\centerline{\includegraphics[width=\picturewidth]{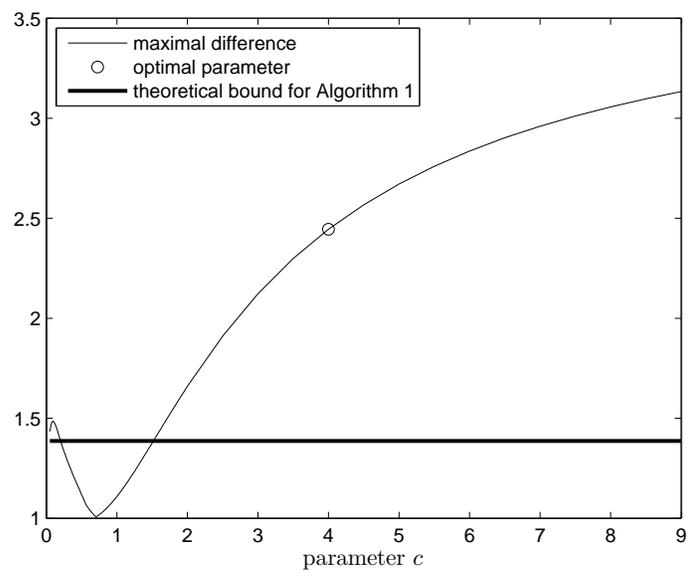}}
\caption{The maximal difference (\ref{eq:maximal-tennis})
  for the WdAA as function of the parameter $c$ on the tennis data.
  The theoretical bound for the WdAA is $5.5452$
  (see Table \ref{tab:football-difference}).}
\label{fig:WdAATennis-all}
\end{center}
\end{figure}

\begin{figure}
\begin{center}
\centerline{\includegraphics[width=\picturewidth]{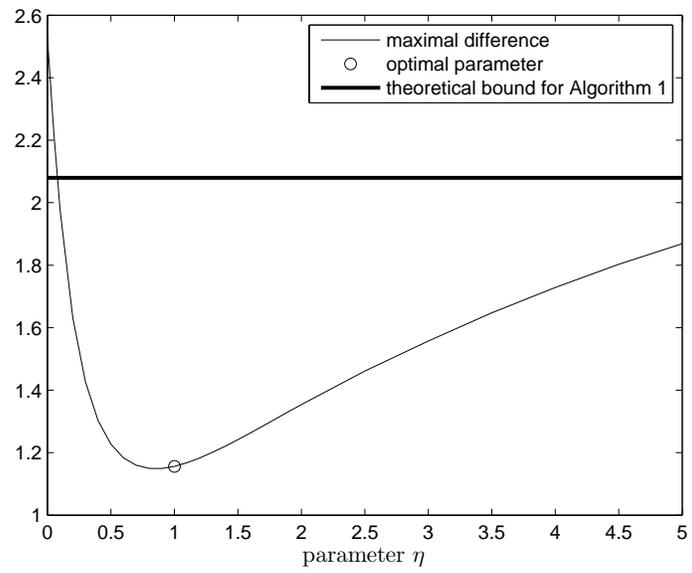}}
\caption{The maximal difference ((\ref{eq:maximal-football}) with $\eta$ in place of $c$)
  for Algorithm~\ref{alg:SAA} as function of the parameter $\eta$ on the football data.}
\label{fig:AAFootball-all}
\end{center}
\end{figure}

\begin{figure}
\begin{center}
\centerline{\includegraphics[width=\picturewidth]{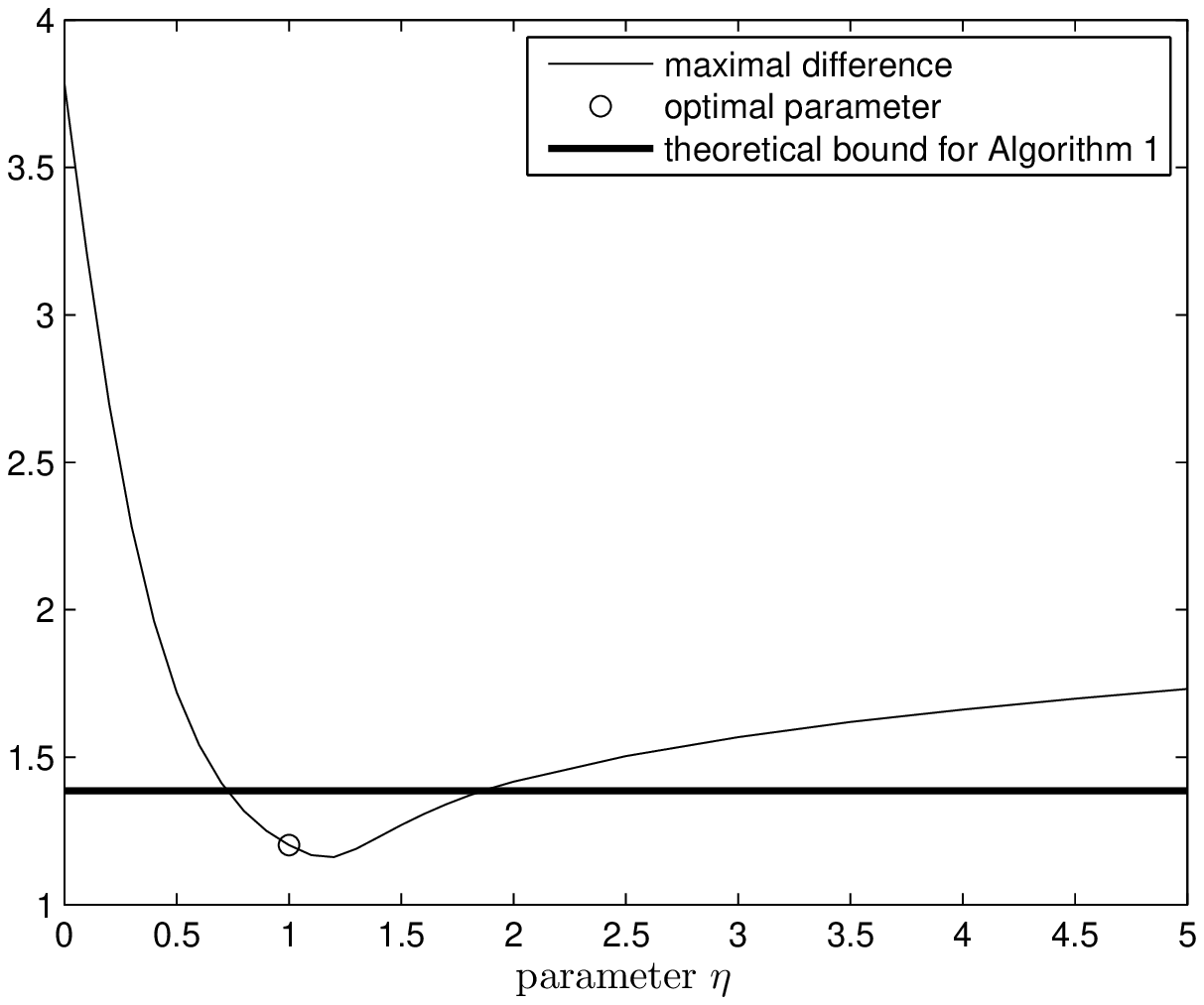}}
\caption{The maximal difference ((\ref{eq:maximal-tennis}) with $\eta$ in place of $c$)
  for Algorithm~\ref{alg:SAA} as function of the parameter $\eta$ on the tennis data.}
\label{fig:AATennis-all}
\end{center}
\end{figure}

Figures \ref{fig:WdAAFootball-1} and \ref{fig:WdAATennis-1}
show the behaviour of the WdAA for the value of parameter $c=1$,
i.e., $\eta=1$,
that is optimal for Algorithm \ref{alg:SAA}.
They look remarkably similar to Figures \ref{fig:football} and \ref{fig:tennis},
respectively.

\begin{figure}
\begin{center}
\centerline{\includegraphics[width=\picturewidth]{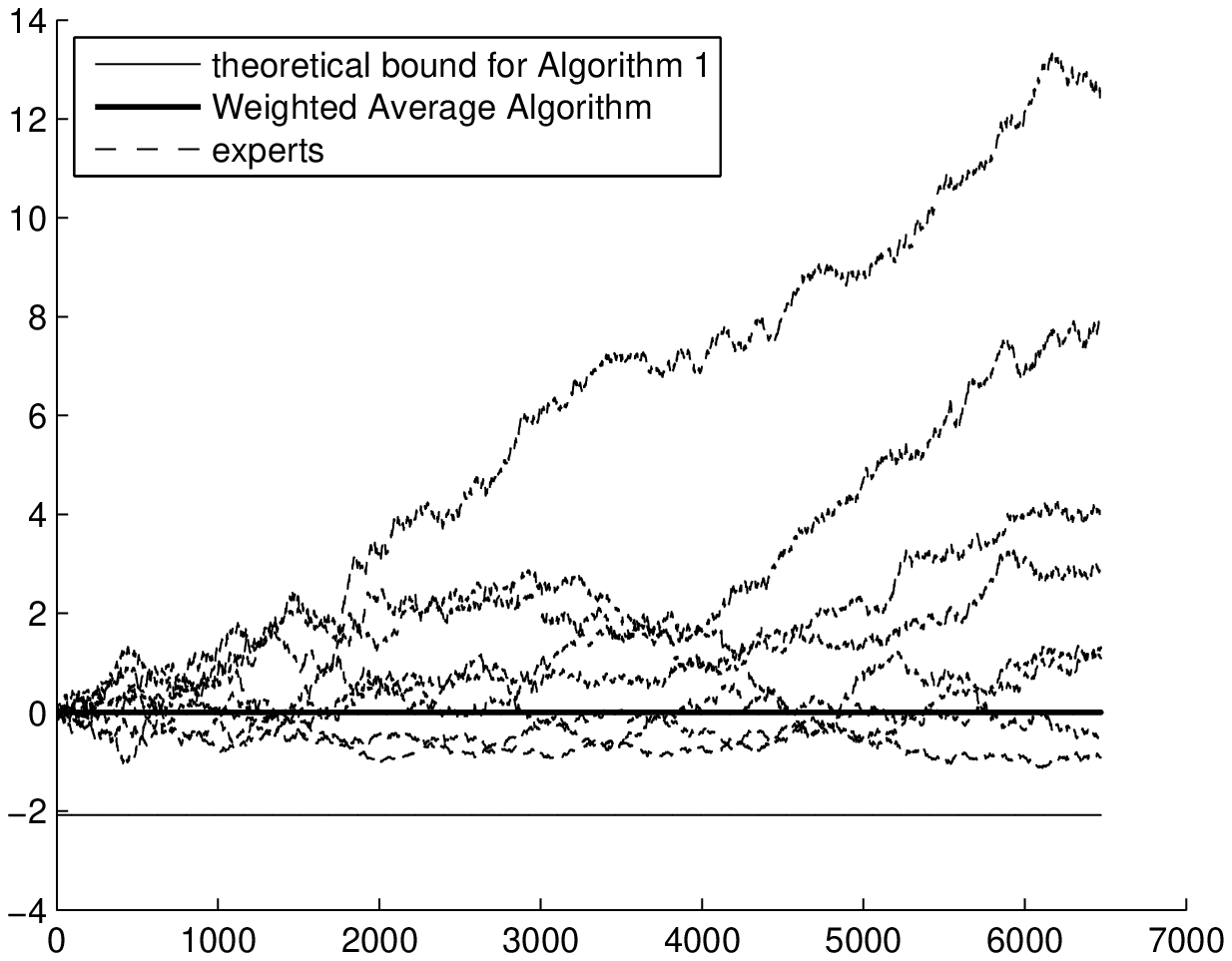}}
\caption{The difference between the cumulative loss of each of the $8$ bookmakers
  and of the WdAA on the football data for $c=1$
  (the value of parameter minimizing the theoretical performance guarantee
  for Algorithm~\ref{alg:SAA}).}
\label{fig:WdAAFootball-1}
\end{center}
\end{figure}

\begin{figure}
\begin{center}
\centerline{\includegraphics[width=\picturewidth]{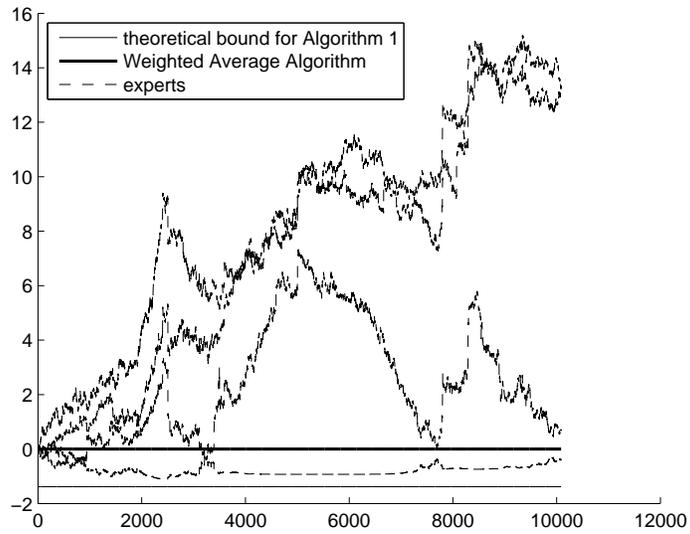}}
\caption{The difference between the cumulative loss of each of the $4$ bookmakers
  and of the WdAA for $c=1$ on the tennis data.}
\label{fig:WdAATennis-1}
\end{center}
\end{figure}

The following two algorithms,
the Weak Aggregating Algorithm (WkAA) and the Hedge algorithm (HA),
make increasingly weaker assumptions about the prediction game being played.
Algorithm \ref{alg:SAA} computes the experts' weights
taking full account of the degree of convexity of the loss function
and uses a minimax optimal substitution function.
Not surprisingly,
it leads to the optimal loss bound of the form (\ref{eq:cannot}).
The WdAA computes the experts' weights in the same way,
but uses a suboptimal substitution function;
this naturally leads to a suboptimal loss bound.
The WkAA ``does not know'' that the loss function is strictly convex;
it computes the experts' weights
in a way that leads to decent results for all convex functions.
The WkAA uses the same substitution function as the WdAA,
but this appears less important than the way it computes the weights.
The HA ``knows'' even less:
it does not even know that its and the experts' performance
is measured using a loss function.
At each step the HA decides which expert it is going to follow,
and at the end of the step it is only told the losses suffered by all experts.
Therefore, it is not surprising that the WkAA
does not perform as well as Algorithm \ref{alg:SAA}
and the WdAA with $c=1$;
the performance of the HA is even weaker:
see Figures \ref{fig:WkAAFootball-all}--\ref{fig:HATennis-all}.
The HA is a randomized algorithm,
so we show the expected performance.

\begin{figure}[p]
\begin{center}
\centerline{\includegraphics[width=\picturewidth]{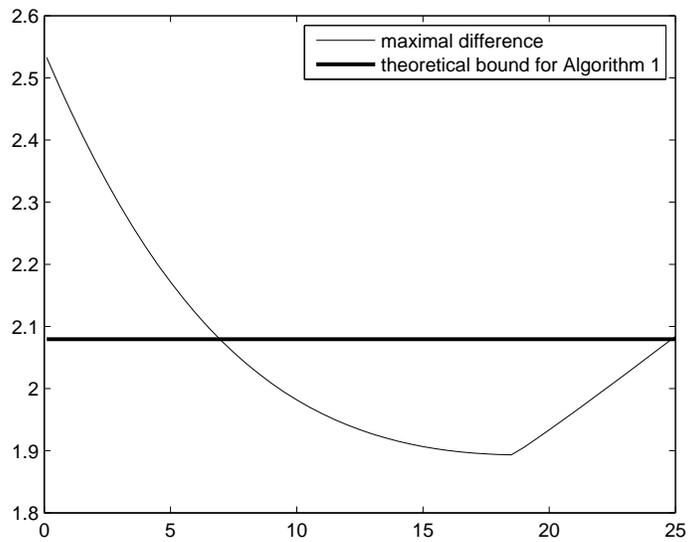}}
\caption{The maximal difference for the Weak Aggregating Algorithm (WkAA)
  as function of $c$ on the football data.}
\label{fig:WkAAFootball-all}
\end{center}
\end{figure}

\begin{figure}[p]
\begin{center}
\centerline{\includegraphics[width=\picturewidth]{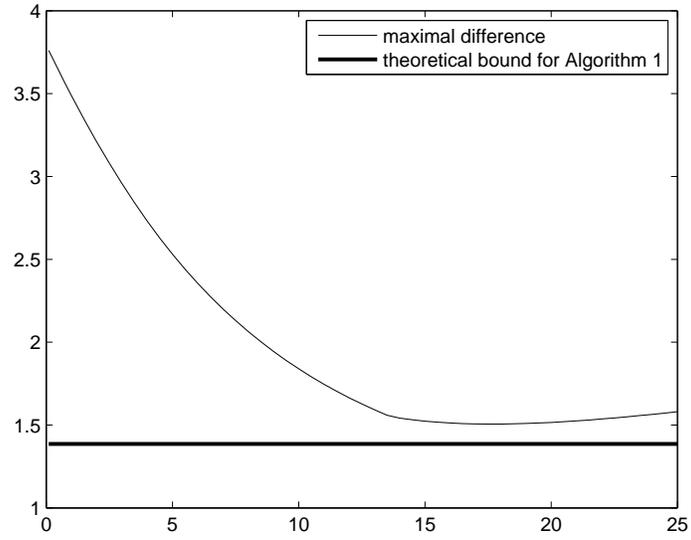}}
\caption{The maximal difference for the WkAA
  as function of $c$ on the tennis data.}
\label{fig:WkAATennis-all}
\end{center}
\end{figure}

\begin{figure}[p]
\begin{center}
\centerline{\includegraphics[width=\picturewidth]{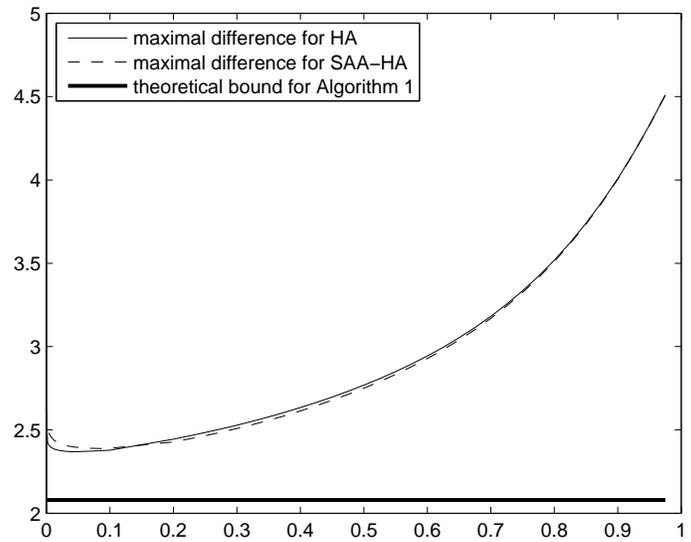}}
\caption{The expected maximal difference for the Hedge algorithm (HA)
  and for the SAA Hedge algorithm (SAA-HA)
  as a function of $\beta$ on the football data.}
\label{fig:HAFootball-all}
\end{center}
\end{figure}

\begin{figure}[p]
\begin{center}
\centerline{\includegraphics[width=\picturewidth]{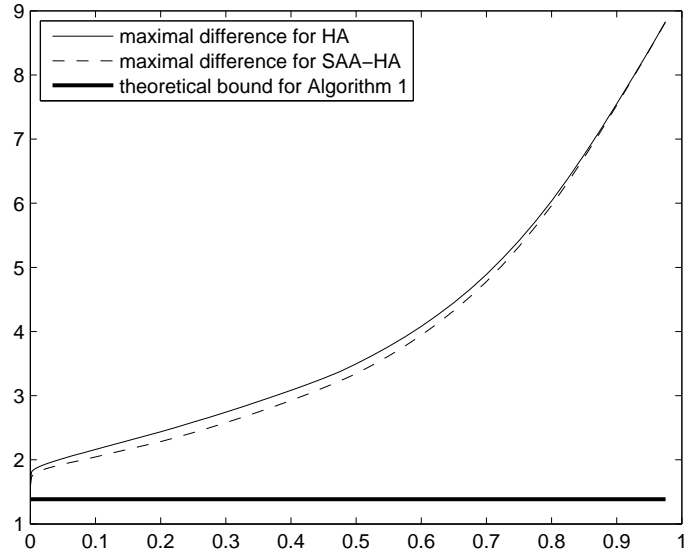}}
\caption{The expected maximal difference for the HA and for the SAA-HA
  as a function of $\beta$ on the tennis data.}
\label{fig:HATennis-all}
\end{center}
\end{figure}

Figures \ref{fig:WkAAFootball-all}--\ref{fig:HATennis-all}
show the performance of the WdAA and the HA
for all possible values of their parameters
($c$ and $\beta$, respectively).
We do not show the optimal values of parameters
since neither algorithm satisfies a loss bound of the form (\ref{eq:cannot})
(typical loss bounds for these algorithms allow $A$ to depend on $N$,
and the optimal value would also depend on $N$).

In the case of the HA,
the loss bound given in the original paper \cite{freund/schapire:1997}
was replaced, in the same framework,
by a stronger bound in \cite{vovk:1998game} (Example 7).
The stronger bound is achieved by the SAA applied to the HA framework
described above (with no loss function);
this algorithm is referred to as SAA-HA in the captions.
The description of the SAA-HA given in \cite{vovk:1998game}
admits some freedom in the choice of Learner's decision;
our implementation replaces the HA's weights $p^k$, $k=1,\ldots,K$,
with
$$
  \frac
  {-\ln\left(1+(\beta-1)p^k\right)}
  {-\sum_{k=1}^K\ln\left(1+(\beta-1)p^k\right)},
  \quad
  k=1,\ldots,K.
$$
The losses suffered by the HA and the SAA-HA are very close.

An interesting observation is that,
for both football and tennis data,
the loss of the HA is almost minimized by setting its parameter $\beta$ to 0
(the qualification ``almost'' is necessary in the case of the tennis data as well:
the lines of maximal difference in Figure~\ref{fig:HATennis-all}
are not monotonic for $\beta$ extremely close to $0$).
The HA with $\beta=0$ coincides with the Follow the Leader Algorithm (FLA),
which chooses the same decision as the best
(with the smallest loss up to now) expert;
if there are several best experts
(which almost never happens after the first step),
their predictions are averaged with equal weights.
Standard examples
(see, e.g., \cite{cesabianchi/lugosi:2006}, Section 4.3)
show that this algorithm
(unlike its version Follow the Perturbed Leader)
can fail badly on some data sequences.
However, its empirical performance
(Figures~\ref{fig:FLAFootball} and~\ref{fig:FLATennis})
on our data sets is not so bad:
it violates the loss bounds for Algorithm \ref{alg:SAA} only slightly.

\begin{figure}[p]
\begin{center}
\centerline{\includegraphics[width=\picturewidth]{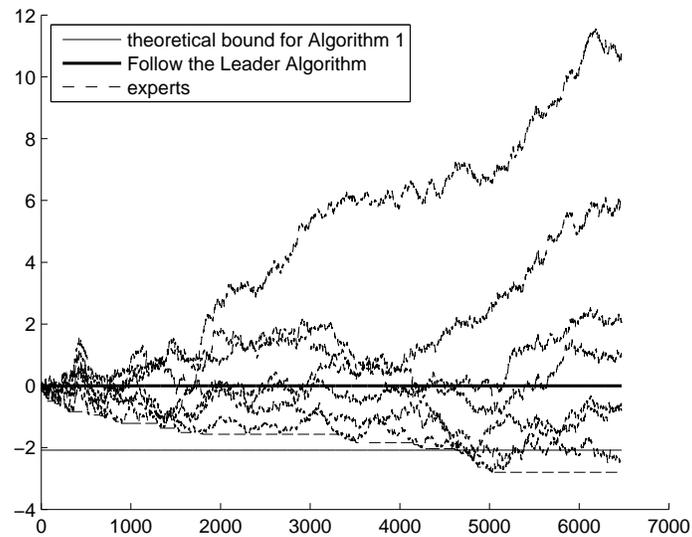}}
\caption{The difference between the cumulative loss of each of the $8$ bookmakers
  and of the Follow the Leader Algorithm on the football data.}
\label{fig:FLAFootball}
\end{center}
\end{figure}

\begin{figure}[p]
\begin{center}
\centerline{\includegraphics[width=\picturewidth]{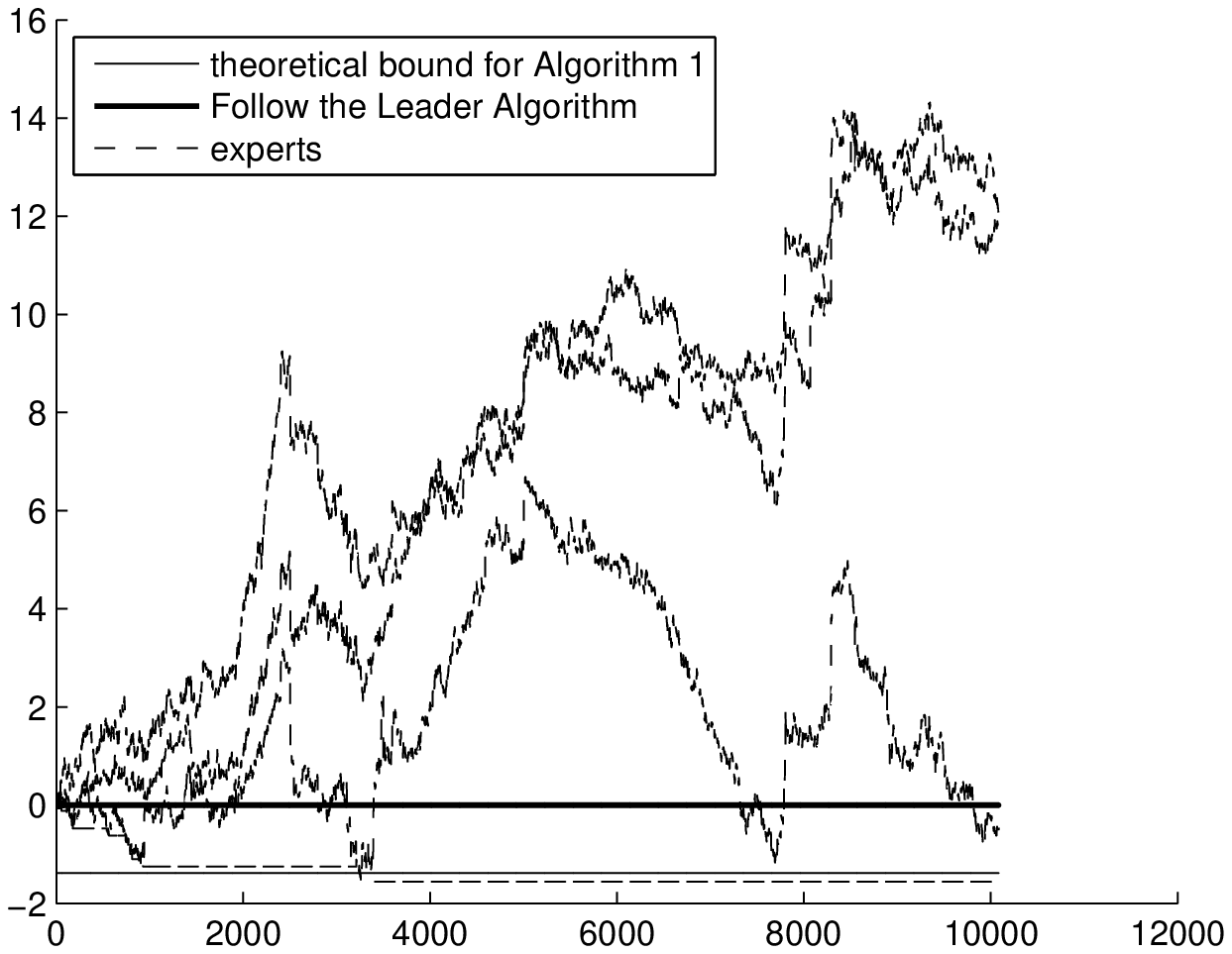}}
\caption{The difference between the cumulative loss of each of the $4$ bookmakers
  and of the Follow the Leader Algorithm on the tennis data.}
\label{fig:FLATennis}
\end{center}
\end{figure}

The decent performance of the Follow the Leader Algorithm suggests
checking the empirical performance of other similarly naive algorithms.
The Simple Average Algorithm's decision
is defined as the arithmetic mean of the experts' decisions
(with equal weights).
Figures \ref{fig:SAAFootball} and \ref{fig:SAATennis}
show the performance of this algorithm.
It does violate the theoretical loss bound for Algorithm \ref{alg:SAA},
but not significantly
(especially in the case of football data).

\begin{figure}[p]
\begin{center}
\centerline{\includegraphics[width=\picturewidth]{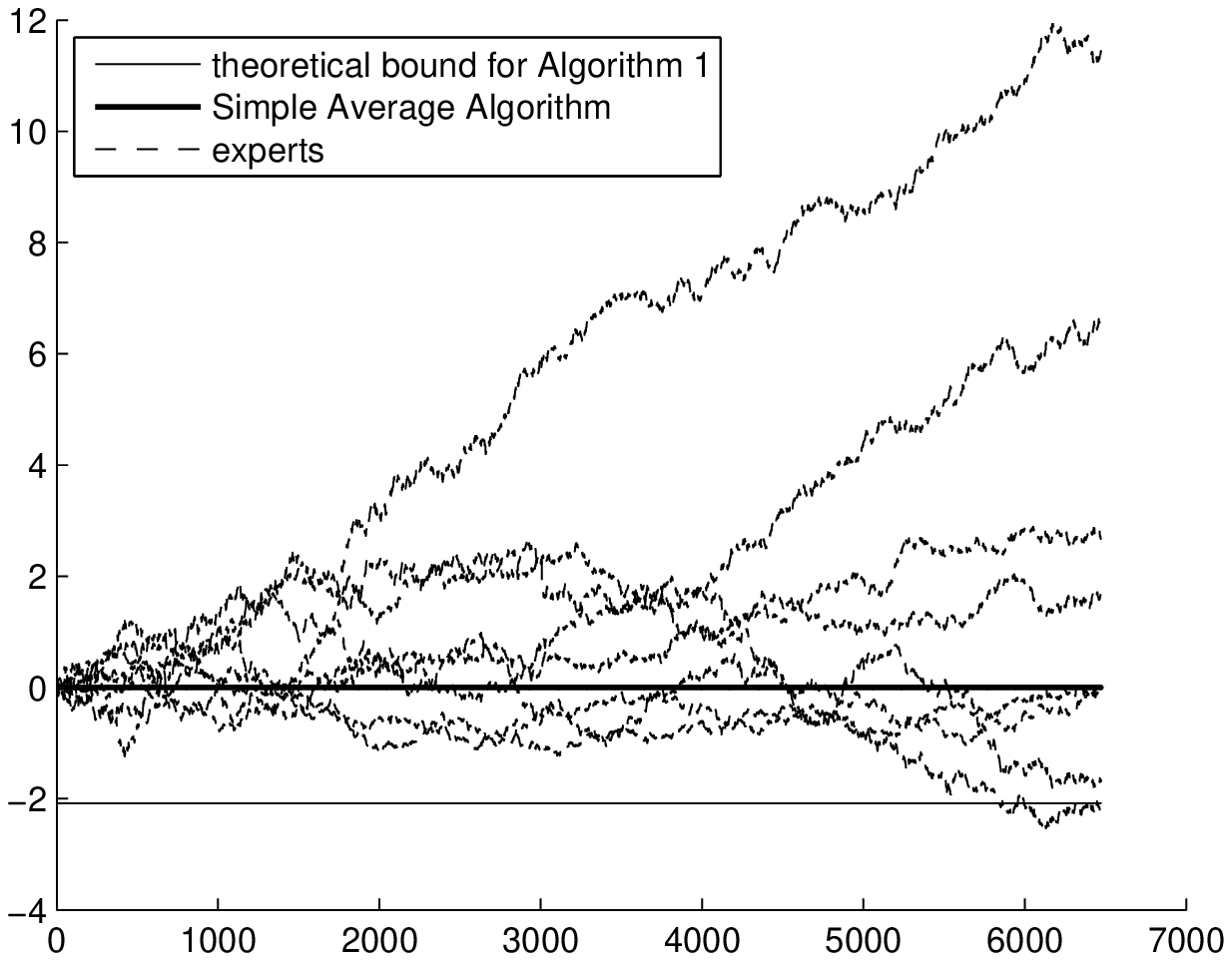}}
\caption{The difference between the cumulative loss of each of the $8$ bookmakers
  and of the Simple Average Algorithm on the football data.}
\label{fig:SAAFootball}
\end{center}
\end{figure}

\begin{figure}[p]
\begin{center}
\centerline{\includegraphics[width=\picturewidth]{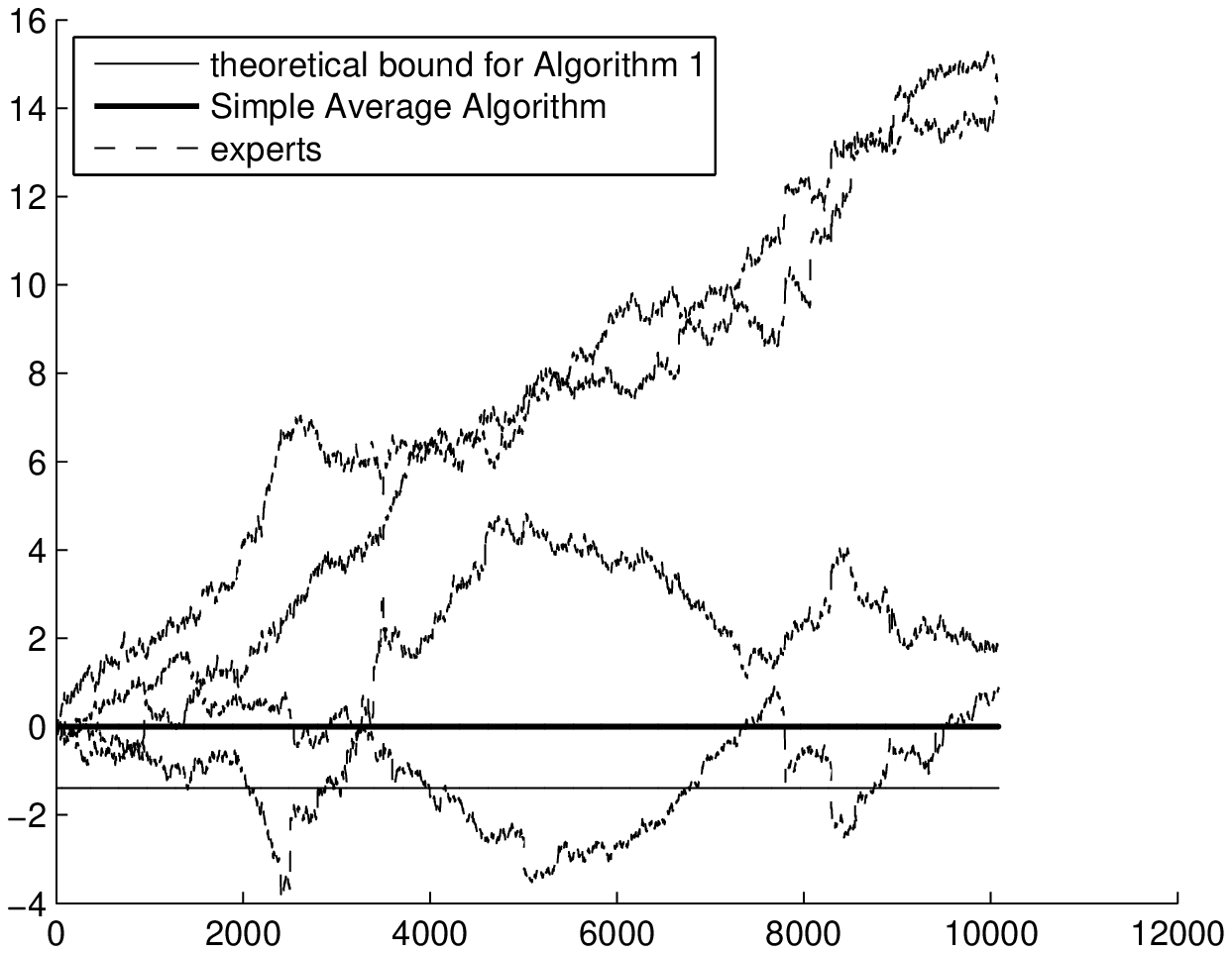}}
\caption{The difference between the cumulative loss of each of the $4$ bookmakers
  and of the Simple Average Algorithm on the tennis data.}
\label{fig:SAATennis}
\end{center}
\end{figure}

The last naive algorithm that we consider
is in fact optimal, but for a different loss function.
The \emph{Bayes Mixture Algorithm} (BMA)
is the Strong Aggregating Algorithm applied
to the log loss function.
This algorithm has a very simple description
\cite{vovk:1990},
and was studied from the point of view of prediction with expert advice
already in \cite{desantis/etal:1988}.
Figures \ref{fig:BMAFootball} and \ref{fig:BMATennis}
show the performance of the BMA
measured by the Brier loss function, as usual.
The performance is excellent for the football data
but much weaker for tennis.

\begin{figure}
\begin{center}
\centerline{\includegraphics[width=\picturewidth]{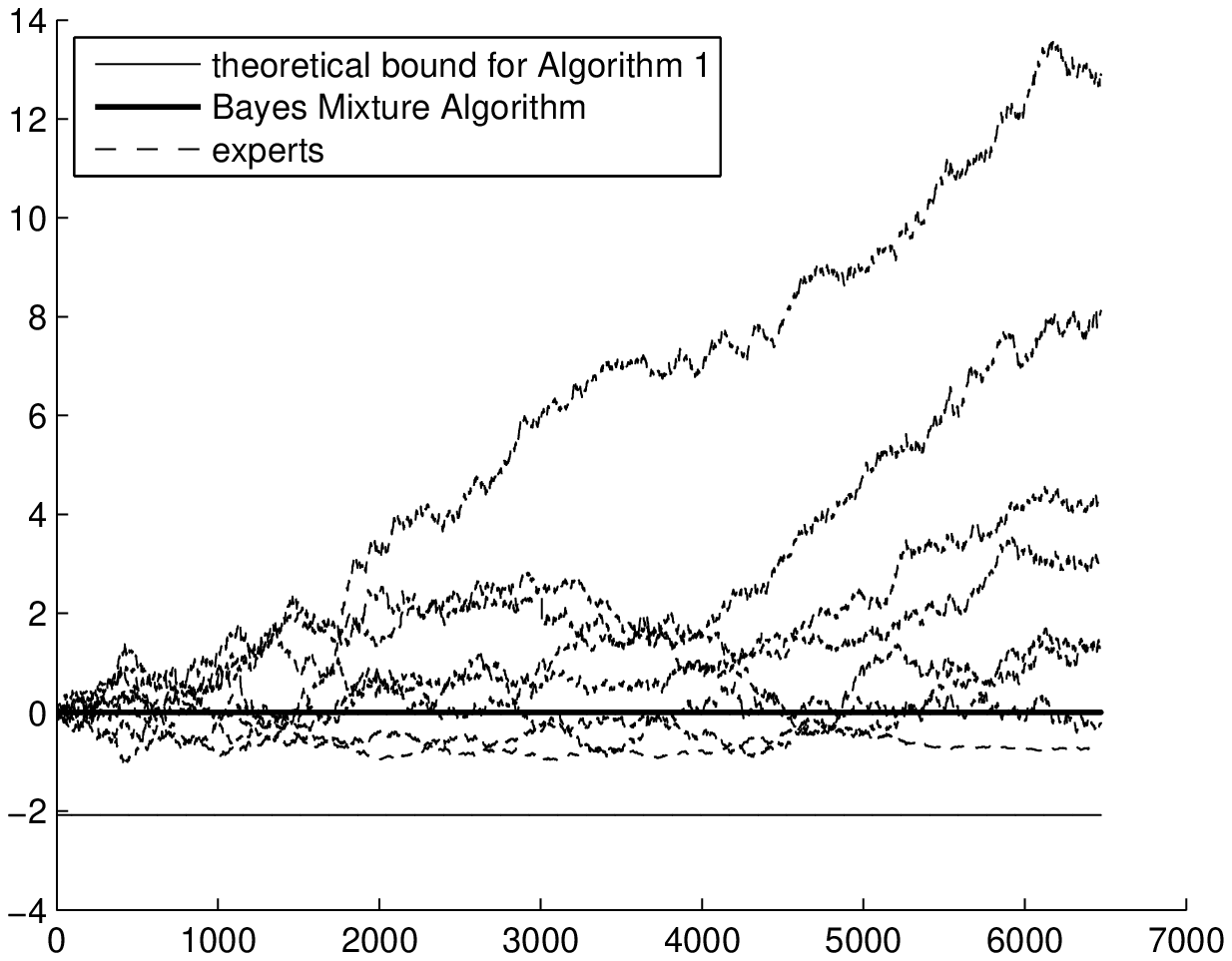}}
\caption{The difference between the cumulative loss of each of the $8$ bookmakers
  and of the Bayes Mixture Algorithm on the football data.}
\label{fig:BMAFootball}
\end{center}
\end{figure}

\begin{figure}
\begin{center}
\centerline{\includegraphics[width=\picturewidth]{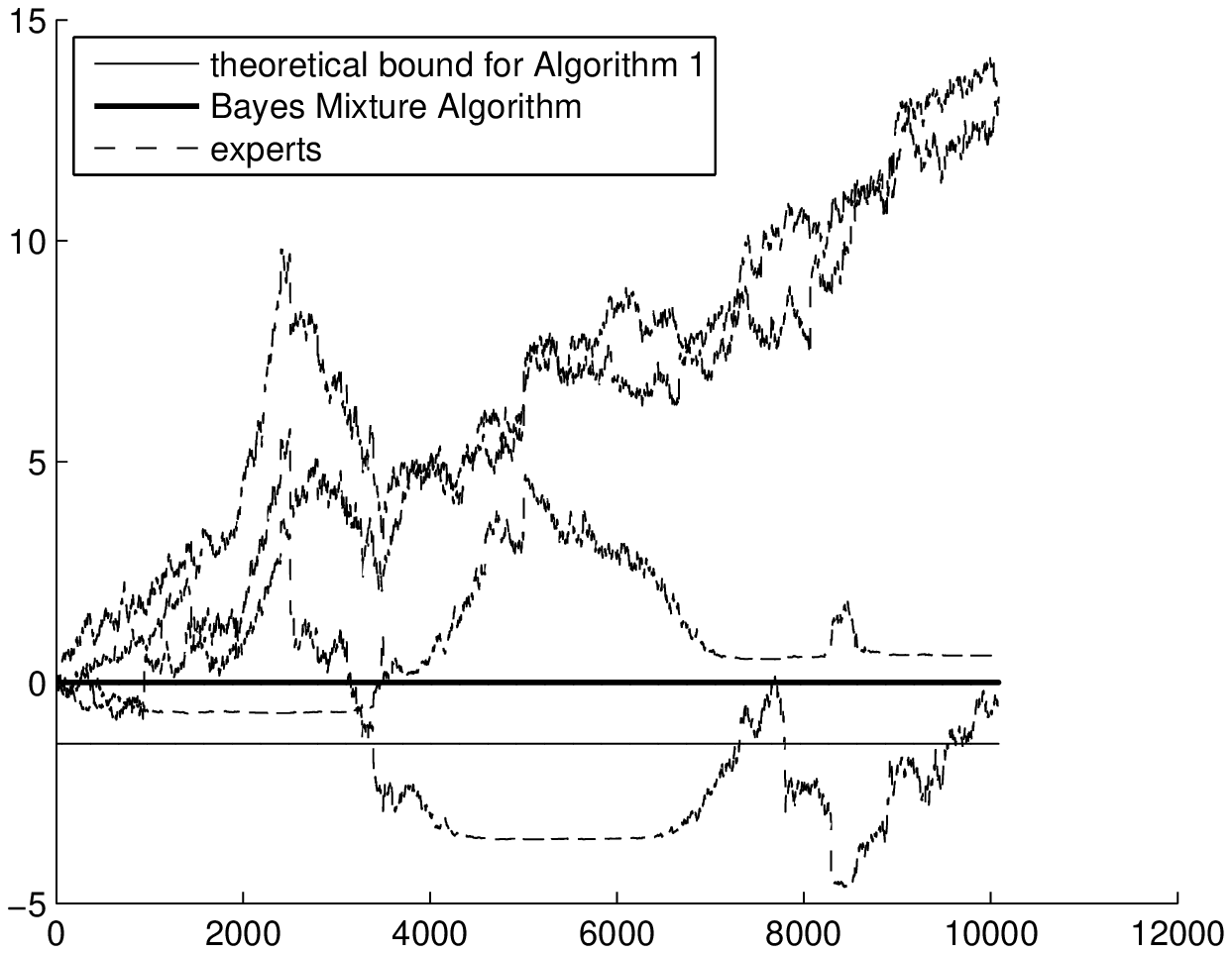}}
\caption{The difference between the cumulative loss of each of the $4$ bookmakers
  and of the Bayes Mixture Algorithm on the tennis data.}
\label{fig:BMATennis}
\end{center}
\end{figure}

Despite the decent performance of the three naive algorithms
on our two data sets,
there is always a danger of catastrophic performance on some data set:
there are no performance guarantees for these algorithms whatsoever.
It is an important advantage of more sophisticated algorithms
that they establish some upper bound on the algorithm's regret.

Precise numbers associated with the figures referred to above
are given in Tables \ref{tab:football-difference}
and \ref{tab:tennis-difference}:
the second column gives the maximal differences
(\ref{eq:maximal-football}) and (\ref{eq:maximal-tennis}),
respectively.
The numbers preceded by ``$\ge$'' are the maximal differences
corresponding to the best value of parameter
chosen in hindsight, after seeing the data set.
Therefore,
the corresponding numbers involve ``data snooping''
and cannot serve as a fair measure of performance.
The third column gives the theoretical performance guarantees
(if available).

\begin{table}
\begin{center}
\begin{tabular}{|c|c|c|}
  \hline
  Algorithm & Maximal difference & Theoretical bound\\
  \hline
  Algorithm \ref{alg:SAA} & 1.1562 & 2.0794\\
  WdAA ($c=16/3$) & 1.6619 & 11.0904\\
  WdAA ($c=1$) & 1.1281 & none of the form (\ref{eq:cannot})\\
  WkAA & ${}\ge1.8933$ & none of the form (\ref{eq:cannot})\\
  HA (expected) & ${}\ge2.3694$ & none of the form (\ref{eq:cannot})\\
  SAA-HA (expected) & ${}\ge2.3882$ & none of the form (\ref{eq:cannot})\\
  Follow the Leader Algorithm & 2.7983 & none\\
  Simple Average Algorithm & 2.5422 & none\\
  Bayes Mixture Algorithm & 1.0602 & none\\
  \hline
\end{tabular}
\end{center}
\caption{The maximal difference between the loss of each algorithm
  and the loss of the best expert for the football data (second column);
  the theoretical upper bound on this difference (third column).}
\label{tab:football-difference}
\end{table}

\begin{table}
\begin{center}
\begin{tabular}{|c|c|c|}
  \hline
  Algorithm & Maximal difference & Theoretical bound\\
  \hline
  Algorithm~\ref{alg:SAA} & 1.2021 & 1.3863\\
  WdAA ($c=4$) & 2.4450 & 5.5452\\
  WdAA ($c=1$) & 1.1089 & none of the form (\ref{eq:cannot})\\
  WkAA & ${}\ge1.5059$ & none of the form (\ref{eq:cannot})\\
  HA (expected) & ${}\ge1.4153$ & none of the form (\ref{eq:cannot})\\
  SAA-HA (expected) & ${}\ge1.3909$ & none of the form (\ref{eq:cannot})\\
  Follow the Leader Algorithm & 1.5597 & none\\
  Simple Average Algorithm & 3.7928 & none\\
  Bayes Mixture Algorithm & 4.6531 & none\\
  \hline
\end{tabular}
\end{center}
\caption{The maximal difference between the loss of each algorithm
  and the loss of the best expert for the tennis data (second column);
  the theoretical upper bound on this difference (third column).}
\label{tab:tennis-difference}
\end{table}
\end{document}